\documentclass[11pt, a4paper, logo]{deepmind}

\usepackage[authoryear, sort&compress, round]{natbib}
\usepackage[noend]{algorithm2e}
\makeatletter
\renewcommand{\@algocf@capt@plain}{above}
\makeatother
\SetAlCapSty{}

\usepackage{color}
\definecolor{deepblue}{rgb}{0,0,0.5}
\definecolor{deepred}{rgb}{0.6,0,0}
\definecolor{deepgreen}{rgb}{0,0.5,0}

\usepackage{dblfloatfix}
\usepackage[normalem]{ulem}
\usepackage{multirow}
\usepackage{microtype}
\usepackage{xspace}
\usepackage{hyperref}

\newcommand\myshade{50}
\colorlet{mylinkcolor}{blue}
\colorlet{mycitecolor}{green}
\colorlet{myurlcolor}{red}
\hypersetup{
  linkcolor  = mylinkcolor!\myshade!black,
  citecolor  = mycitecolor!\myshade!black,
  urlcolor   = myurlcolor!\myshade!black,
  colorlinks = true,
}

\makeatletter
\addto\extrasenglish{%
\renewcommand{\sectionautorefname}{\S\@gobble}%
\renewcommand{\subsectionautorefname}{\S\@gobble}%
\renewcommand{\subsubsectionautorefname}{\S\@gobble}%
}
\makeatother

\usepackage{listings}

\usepackage{courier}
\newcommand{\RR}{\mathbb{R}}
\newcommand{\VV}{\mathbb{V}}

\newcommand{\Cc}{\mathcal{C}}
\newcommand{\Dd}{\mathcal{D}}

\newcommand{\Oo}{\mathcal{O}}

\renewcommand{\phi}{\varphi}

\renewcommand{\leq}{\leqslant}
\renewcommand{\geq}{\geqslant}

\renewcommand{\epsilon}{\varepsilon}
\renewcommand{\imath}{\mathrm{i}}

\newlength{\restsubwidth}
\newlength{\restsubheight}
\newlength{\restsubmoreheight}
\newcommand{\rest}[2]{%
        \settowidth{\restsubwidth}{\ensuremath{#2}}
        \settoheight{\restsubheight}{\ensuremath{{}_{#2}}}
        \ensuremath{{#1\hskip 0.5pt}_{\vrule\kern2pt\parbox[b][%
        4pt][b]{\the\restsubwidth}{%
                        \ensuremath{{}_{#2}}}}}
        }

\newcommand{\lstm}{\textsc{LSTM}}
\newcommand{\ffw} {\textsc{Ffw}}
\newcommand{\attn}{\textsc{Attn}}
\newcommand{\encoder} {\textsc{Encoder}}

\newcommand{\ccca}{\textsc{Cca}}
\newcommand{\crossattention}{\textsc{Ca}}
\newcommand{\softmax}{\textrm{softmax}}
\newcommand{\ret}{\textsc{Ret}}

\newcommand{\MassiveText} {Massive Text}
\newcommand{\bert}{\textsc{Bert}\xspace}

\renewcommand{\encoder}{\textsc{Encoder}}
\newcommand{\knnlm}{k\textrm{NN-LM}}
\newcommand{\knn}{k\textrm{NN}}
\newcommand{\spalm}{\textsc{Spalm}}
\newcommand{\dpr}{\textsc{Dpr}}
\newcommand{\rag}{\textsc{RAG}}
\newcommand{\realm}{\textsc{Realm}}
\newcommand{\fid}{\textsc{FiD}}
\newcommand{\emdr}{\textsc{Emdr}^2}
\newcommand{\retro}{\textsc{Retro}\xspace}
\newcommand{\retroon}{\textsc{Retro[On]}\xspace}
\newcommand{\retrooff}{\textsc{Retro[Off]}\xspace}
\newcommand{\retrofit}{\textsc{Retro}fit\xspace}
\newcommand{\retrofitting}{\textsc{Retro}fitting\xspace}
\newcommand{\lm}{\textsc{Lm}}

\newcommand{\emb}{\textsc{Emb}}
\newcommand{\readout}{\textsc{Read}}
\newcommand{\spli}{\textsc{Split}}
\newcommand{\bpb}{\textrm{bpb}}
\newcommand{\dmodel}{d}
\newcommand{\dffw}{d_{\textrm{ffw}}}
\newcommand{\lcp}{\textrm{LCP}}

\SetCommentSty{mycommfont}

\definecolor{cm_0}{rgb}{0.267,0.005,0.329}
\definecolor{cm_1}{rgb}{0.273,0.026,0.353}
\definecolor{cm_2}{rgb}{0.277,0.050,0.376}
\definecolor{cm_3}{rgb}{0.280,0.073,0.397}
\definecolor{cm_4}{rgb}{0.282,0.095,0.417}
\definecolor{cm_5}{rgb}{0.283,0.116,0.436}
\definecolor{cm_6}{rgb}{0.283,0.136,0.453}
\definecolor{cm_7}{rgb}{0.281,0.156,0.469}
\definecolor{cm_8}{rgb}{0.279,0.175,0.483}
\definecolor{cm_9}{rgb}{0.275,0.195,0.496}
\definecolor{cm_10}{rgb}{0.271,0.214,0.507}
\definecolor{cm_11}{rgb}{0.265,0.233,0.517}
\definecolor{cm_12}{rgb}{0.259,0.252,0.525}
\definecolor{cm_13}{rgb}{0.252,0.270,0.532}
\definecolor{cm_14}{rgb}{0.245,0.288,0.537}
\definecolor{cm_15}{rgb}{0.237,0.305,0.542}
\definecolor{cm_16}{rgb}{0.230,0.322,0.546}
\definecolor{cm_17}{rgb}{0.222,0.339,0.549}
\definecolor{cm_18}{rgb}{0.214,0.356,0.551}
\definecolor{cm_19}{rgb}{0.207,0.372,0.553}
\definecolor{cm_20}{rgb}{0.199,0.388,0.555}
\definecolor{cm_21}{rgb}{0.192,0.403,0.556}
\definecolor{cm_22}{rgb}{0.186,0.419,0.557}
\definecolor{cm_23}{rgb}{0.179,0.434,0.557}
\definecolor{cm_24}{rgb}{0.173,0.449,0.558}
\definecolor{cm_25}{rgb}{0.167,0.464,0.558}
\definecolor{cm_26}{rgb}{0.161,0.479,0.558}
\definecolor{cm_27}{rgb}{0.155,0.493,0.558}
\definecolor{cm_28}{rgb}{0.149,0.508,0.557}
\definecolor{cm_29}{rgb}{0.143,0.523,0.556}
\definecolor{cm_30}{rgb}{0.138,0.537,0.555}
\definecolor{cm_31}{rgb}{0.132,0.552,0.553}
\definecolor{cm_32}{rgb}{0.128,0.567,0.551}
\definecolor{cm_33}{rgb}{0.123,0.582,0.547}
\definecolor{cm_34}{rgb}{0.121,0.596,0.544}
\definecolor{cm_35}{rgb}{0.119,0.611,0.539}
\definecolor{cm_36}{rgb}{0.121,0.626,0.533}
\definecolor{cm_37}{rgb}{0.125,0.640,0.527}
\definecolor{cm_38}{rgb}{0.132,0.655,0.520}
\definecolor{cm_39}{rgb}{0.143,0.669,0.511}
\definecolor{cm_40}{rgb}{0.158,0.684,0.502}
\definecolor{cm_41}{rgb}{0.176,0.698,0.491}
\definecolor{cm_42}{rgb}{0.197,0.712,0.479}
\definecolor{cm_43}{rgb}{0.220,0.726,0.466}
\definecolor{cm_44}{rgb}{0.246,0.739,0.452}
\definecolor{cm_45}{rgb}{0.274,0.752,0.437}
\definecolor{cm_46}{rgb}{0.304,0.765,0.420}
\definecolor{cm_47}{rgb}{0.336,0.777,0.402}
\definecolor{cm_48}{rgb}{0.369,0.789,0.383}
\definecolor{cm_49}{rgb}{0.404,0.800,0.363}
\definecolor{cm_50}{rgb}{0.440,0.811,0.341}
\definecolor{cm_51}{rgb}{0.478,0.821,0.318}
\definecolor{cm_52}{rgb}{0.516,0.831,0.294}
\definecolor{cm_53}{rgb}{0.555,0.840,0.269}
\definecolor{cm_54}{rgb}{0.596,0.849,0.243}
\definecolor{cm_55}{rgb}{0.637,0.857,0.217}
\definecolor{cm_56}{rgb}{0.678,0.864,0.190}
\definecolor{cm_57}{rgb}{0.720,0.870,0.163}
\definecolor{cm_58}{rgb}{0.762,0.876,0.137}
\definecolor{cm_59}{rgb}{0.804,0.882,0.115}
\definecolor{cm_60}{rgb}{0.846,0.887,0.100}
\definecolor{cm_61}{rgb}{0.886,0.892,0.095}
\definecolor{cm_62}{rgb}{0.926,0.897,0.104}
\definecolor{cm_63}{rgb}{0.965,0.902,0.124}

\newcommand{\datavisheader}[0]{
$C_u$ colored by loss difference & $C_u$ colored by LCP with $\ret(C_u{-1})$  & $[N_u^1, F_u^1]$ colored by LCP with $C_{u+1}$ & $[N_u^2, F_u^2]$ colored by LCP with $C_{u+1}$\\ 
$L_{\textsc{\retrooff}} - L_{\retro}\colorbox{cm_0!40}{$ \leq -0.5$}, \colorbox{cm_31!40}{$=0 $},\colorbox{cm_63!40}{$ \geq 0.5$}$
& $ \lcp = $ \colorbox{cm_0!40}{$0$}, \colorbox{cm_12!40}{$1 $}, \colorbox{cm_25!40}{$2 $}, \colorbox{cm_38!40}{$3 $},\colorbox{cm_51!40}{$4 $},\colorbox{cm_63!40}{$ \geq 5 $} & $ \lcp = $  \colorbox{cm_0!40}{$0$}, \colorbox{cm_12!40}{$1 $}, \colorbox{cm_25!40}{$2 $}, \colorbox{cm_38!40}{$3 $},\colorbox{cm_51!40}{$4 $},\colorbox{cm_63!40}{$ \geq 5 $} &
$ \lcp = $ \colorbox{cm_0!40}{$0$}, \colorbox{cm_12!40}{$1 $}, \colorbox{cm_25!40}{$2 $}, \colorbox{cm_38!40}{$3 $},\colorbox{cm_51!40}{$4 $},\colorbox{cm_63!40}{$ \geq 5 $} \\
\toprule}

\newcommand{\samplevisheader}[0]{
Prompt and sample of $\retrooff$  & Prompt and sample of $\retroon$ & $[N_u^1, F_u^1]$ colored by LCP with $C_{u+1}$ & $[N_u^2, F_u^2]$ colored by LCP with $C_{u+1}$\\ 

& 
colored by LCP with $\ret(C_u{-1})$ 
& & \\

&$ \lcp = $ \colorbox{cm_0!40}{$0$}, \colorbox{cm_12!40}{$1 $}, \colorbox{cm_25!40}{$2 $}, \colorbox{cm_38!40}{$3 $},\colorbox{cm_51!40}{$4 $},\colorbox{cm_63!40}{$ \geq 5 $} & $ \lcp = $  \colorbox{cm_0!40}{$0$}, \colorbox{cm_12!40}{$1 $}, \colorbox{cm_25!40}{$2 $}, \colorbox{cm_38!40}{$3 $},\colorbox{cm_51!40}{$4 $},\colorbox{cm_63!40}{$ \geq 5 $}  &$ \lcp = $ \colorbox{cm_0!40}{$0$}, \colorbox{cm_12!40}{$1 $}, \colorbox{cm_25!40}{$2 $}, \colorbox{cm_38!40}{$3 $},\colorbox{cm_51!40}{$4 $},\colorbox{cm_63!40}{$ \geq 5 $} \\ 

\toprule}

\usepackage[utf8]{inputenc}
\usepackage{newunicodechar}

\usepackage{CJKutf8}

\title{Improving language models by retrieving from trillions of tokens}

\correspondingauthor{ \{sborgeaud|amensch|jordanhoffmann|sifre\}@deepmind.com}

\paperurl{}

\reportnumber{} 

\renewcommand{\today}{}

\author[$\dag$]{Sebastian~Borgeaud}
\author[$\dag$]{Arthur~Mensch}
\author[$\dag$]{Jordan~Hoffmann}
\author[  \hspace{-.8ex}]{Trevor~Cai}
\author[  \hspace{-.8ex}]{Eliza~Rutherford}
\author[  \hspace{-.8ex}]{Katie~Millican}
\author[  \hspace{-.8ex}]{George~van~den~Driessche}
\author[  \hspace{-.8ex}]{Jean-Baptiste~Lespiau}
\author[  \hspace{-.8ex}]{Bogdan~Damoc}
\author[  \hspace{-.8ex}]{Aidan~Clark}
\author[  \hspace{-.8ex}]{Diego~de~Las~Casas}
\author[  \hspace{-.8ex}]{Aurelia~Guy}
\author[  \hspace{-.8ex}]{Jacob~Menick}
\author[  \hspace{-.8ex}]{Roman~Ring}
\author[  \hspace{-.8ex}]{Tom~Hennigan}
\author[  \hspace{-.8ex}]{Saffron~Huang}
\author[  \hspace{-.8ex}]{Loren~Maggiore}
\author[  \hspace{-.8ex}]{Chris~Jones}
\author[  \hspace{-.8ex}]{Albin~Cassirer}
\author[  \hspace{-.8ex}]{Andy~Brock}
\author[  \hspace{-.8ex}]{Michela~Paganini}
\author[  \hspace{-.8ex}]{Geoffrey~Irving}
\author[  \hspace{-.8ex}]{Oriol~Vinyals}
\author[  \hspace{-.8ex}]{Simon~Osindero}
\author[  \hspace{-.8ex}]{Karen~Simonyan}
\author[$\ddag$]{Jack~W.~Rae}
\author[$\ddag$]{Erich~Elsen}
\author[$\dag$,$\ddag$]{Laurent~Sifre}

\affil[ ]{All authors from DeepMind}
\affil[$\dag$]{Equal contributions}
\affil[$\ddag$]{Equal senior authorship}

\begin{abstract}
We enhance auto-regressive language models by conditioning on document chunks retrieved from a large corpus, based on local similarity with preceding tokens.
With a 2 trillion token database, our Retrieval-Enhanced Transformer ($\retro$) obtains comparable performance to GPT-3 and Jurassic-1 on the~Pile, despite using 25$\times$ fewer parameters. 
After fine-tuning, $\retro$ performance translates to downstream knowledge-intensive tasks such as question answering.
$\retro$ combines a frozen $\bert$ retriever, a differentiable encoder and a chunked cross-attention mechanism to predict tokens based on an order of magnitude more data than what is typically consumed during training.
We typically train $\retro$ from scratch, yet can also rapidly $\retro$fit pre-trained transformers with retrieval and still achieve good performance. 
Our work opens up new avenues for improving language models through explicit memory at unprecedented scale.
\end{abstract}

\begin{document}

\maketitle
\section{Introduction}
Language modelling (LM) is an unsupervised task that consists of modelling the probability of text, usually by factorising it into conditional next-token predictions $p(x_1, \dotsc, x_n) = \prod_i p(x_i | x_{<i})$. 
Neural networks have proven to be powerful language models, first in the form of recurrent architectures \citep{mikolov2010recurrent, graves2013generating, jozefowicz2016exploring} and more recently in the form of Transformers \citep{vaswani2017attention}, that use attention to contextualise the past.
Large performance improvements have come from increasing the amount of data, training compute, or model parameters.
Transformers have been scaled from $100$ million parameter models in seminal work to over hundred billion parameters \citep{radford2019language, brown2020language} in the last two years which has led to models that do very well on a wide array of tasks in a zero or few-shot formulation. Increasing model size predictably improves performance on a wide range of downstream tasks \citep{scaling_laws}. 
The benefits of increasing the number of parameters come from two factors: additional computations at training and inference time, and increased memorization of the training data.

In this work, we endeavor to decouple these, by exploring efficient means of augmenting language models with a massive-scale memory without significantly increasing computations. 
Specifically, we suggest retrieval from a large text database as a complementary path to scaling language models. Instead of increasing the size of the model and training on more data, we equip models with the ability to directly access a large database to perform predictions---a semi-parametric approach.
At a high level, our Retrieval Transformer ($\retro$) model splits the input sequence into chunks and retrieves text similar to the previous chunk to improve the predictions in the current chunk.
Existing retrieval for language modelling work only considers small transformers ($100$ millions parameters) and databases of limited size (up to billions of tokens) \citep{khandelwal2020generalization, yogatama2021adaptive, guu2020retrieval, lewis2020retrieval}. 
To our knowledge, our work is the first to show the benefits of scaling the retrieval database to trillions of tokens for large parametric language models. Our main contributions are the following.

\begin{figure}[t]
    \centering
    \includegraphics[width=.95\textwidth]{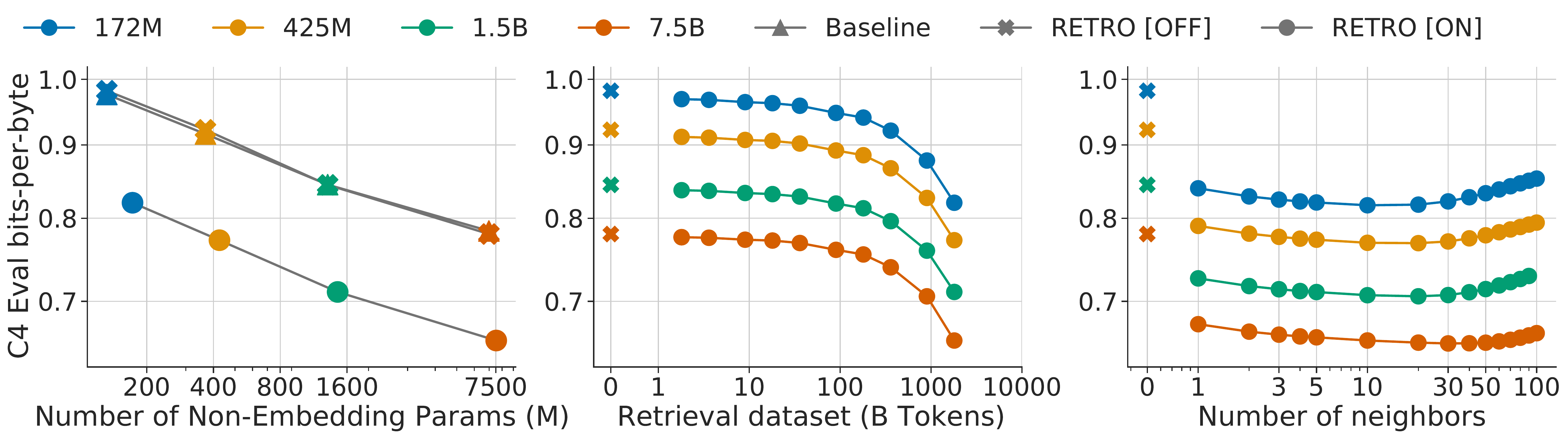}
    \caption{\textbf{Scaling of $\retro$.} 
    The performance gain of our retrieval models remains constant with model scale (left), and is comparable to multiplying the parameteric model size by $\sim 10 \times$. The gain increases with the size of the retrieval database (middle) and the number of retrieved neighbours (right) on the C4 validation set, when using up to 40 neighbours. Past this, performance begins to degrade, perhaps due to the reduced quality. At evaluation $\retro$ can be used without retrieval data ($\retro$[OFF]), bringing limited performance degradation compared to baseline transformers.}
    \label{fig:summary}
\end{figure}

\begin{itemize}
    \item We introduce $\retro$, a retrieval-enhanced autoregressive language model (\autoref{subsec:probabilistic-model}). We use a chunked cross-attention module to incorporate the retrieved text (\autoref{subsec-model}), with time complexity linear in the amount of retrieved data. We show that retrieving based on a pre-trained frozen $\bert$ model (\autoref{subsec-data-preparation}) works at scale, removing the need for training and updating a retriever network.
    
    \item We show that our method scales well with model size and database size (\autoref{fig:summary}): $\retro$ provides a constant gain for models ranging from 150M to 7B parameters, and $\retro$ can be improved at evaluation time by increasing the database size and the number of retrieved neighbours. Our largest model obtains state-of-the-art results on a range of downstream evaluation datasets including Wikitext103 \citep{merity2016pointer} and the Pile \citep{pile} (\autoref{sec-results}). We show that \retro can be fine-tuned to achieve competitive performance on downstream tasks such as question answering (\autoref{subsection-qa}).
    
    \item We propose an evaluation aware of proximity of test documents with the training set (\autoref{sec:leakage-quantification}), addressing the problem of test set leakage \citep{lee2021deduplicating}. This is relevant for all language models, and especially for retrieval-enhanced models since they have direct access to the training dataset during evaluation.
    Using this methodology, we show that the performance of $\retro$ comes from both explicit neighbour copying and general knowledge extraction (\autoref{subsec-analysis-train-valid-leak}).
\end{itemize}

\section{Method}
\label{sec-methods}

\begin{figure}[t]
    \centering
    \includegraphics[width=1.\textwidth]{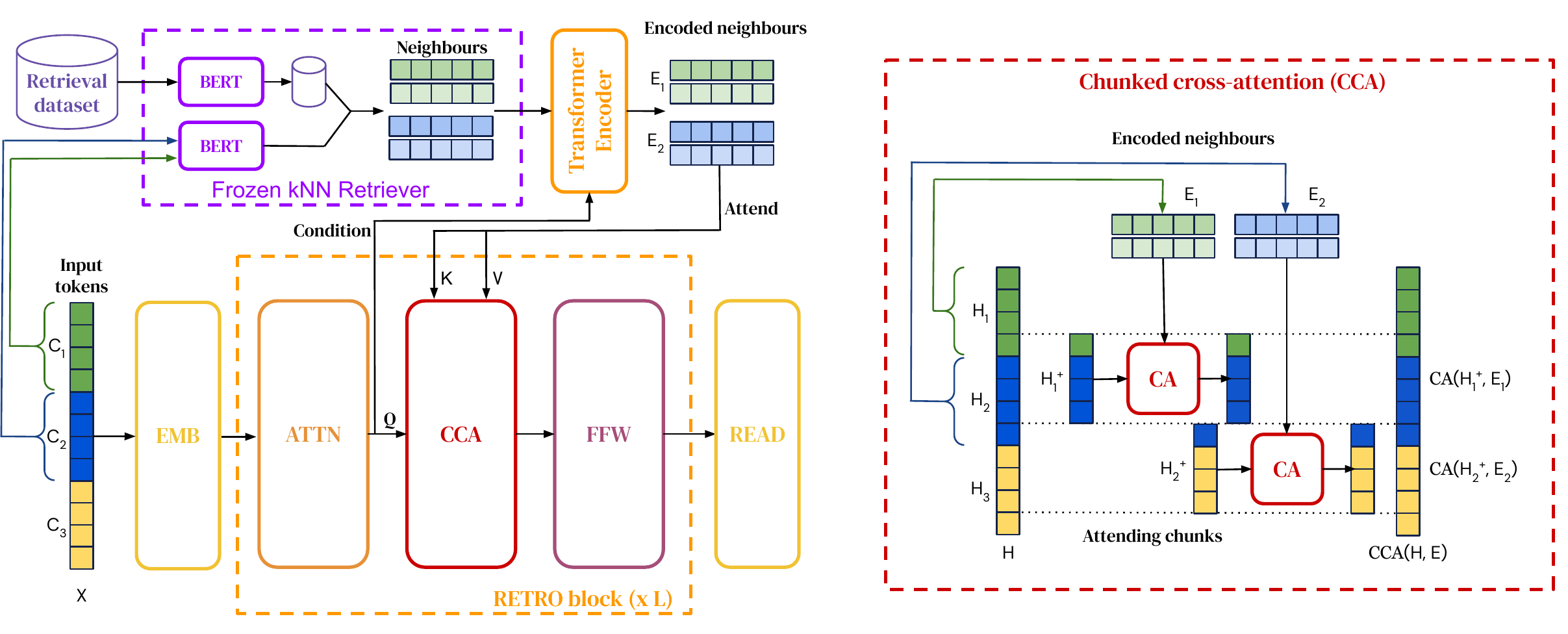}
    \caption{\textbf{$\retro$ architecture.} \textit{Left:} simplified version where a sequence of length $n=12$ is split into $l=3$ chunks of size $m=4$. For each chunk, we retrieve $k=2$ neighbours of $r=5$ tokens each. The retrieval pathway is shown on top. \textit{Right:} Details of the interactions in the $\ccca$ operator. Causality is maintained as neighbours of the first chunk only affect the last token of the first chunk and tokens from the second chunk.}
    \label{fig:architecture-summary}
\end{figure}

We design our retrieval-enhanced architecture to be capable of retrieving from a database with trillions of tokens. 
For this purpose, we retrieve at the level of contiguous token \textit{chunks} instead of individual tokens which reduces storage and computation requirements by a large linear factor.
Our method first constructs a key-value database, where values store raw chunks of text tokens and keys are frozen $\bert$ embedddings~\citep{devlin2019bert}. We use a frozen model to avoid having to periodically re-compute embeddings over the entire database during training.
Each training sequence is then split into chunks, which are augmented with their $k$-nearest neighbour retrieved from the database. An encoder-decoder architecture integrates retrieval chunks into the model's predictions.
We summarize the $\retro$ architecture in \autoref{fig:architecture-summary}, and detail it in this section. We end the section by introducing a new methodology to evaluate language models when an evaluation set is partially present in the training set.

\subsection{Training dataset}
\label{subsec-training-dataset}
We use a multi-lingual version of \textit{MassiveText} \citep{rae2021gopher} for both training and retrieval data.
The dataset consists of text documents from multiple sources and multiple languages totalling over 5 trillion tokens (detailed in \autoref{table-massive-text}). Sequences are sampled from subsets of the training data, with sampling weights given in the right-most column of \autoref{table-massive-text}.
We tokenize the dataset using SentencePiece \citep{kudo2018} with a vocabulary of 128,000 tokens.
During training (unless otherwise specified), we retrieve from 600B tokens from the training data. The training retrieval database is made of the same subsets as the training data, in proportion that matches the training sampling frequencies. 
During evaluation the retrieval database consists in the full union of these datasets, with the exception of books for which we use a sub-sample of 4\%.
The evaluation retrieval database thus contains 1.75T tokens.
To limit test set leakage, we compute the $13$-gram Jaccard similarity between train and test documents using the MinHash scheme and remove all training documents with high similarity (0.8 or higher) to a validation or test set document. 
Additionally, we remove all validation and test articles from Wikitext103 \citep{merity2016pointer} from our Wikipedia training data.

\begin{table}[b]
\caption{\textbf{MassiveText}. The last column indicates the sampling weight during training. The multilingual subsets include documents in 10 languages. The full breakdown is given in \autoref{appendix:massive-text}.}
\small{
\begin{center}
\begin{tabular}{c r c c c}
Source &  Token count (M) & Documents (M) & Multilingual & Sampling frequency \\ [0.5ex] 
\toprule
Web & 977,563 & 1,208 & Yes & 55\% \\
Books &  3,423,740 & 20 & No & 25\% \\
News & 236,918 & 398 & No & 10\% \\
Wikipedia & 13,288 & 23 & Yes & 5\% \\
GitHub & 374,952 & 143 & No & 5\% \\
\end{tabular}
\end{center}
}
\label{table-massive-text}
\end{table}

\subsection{Retrieval-enhanced autoregressive token models}\label{subsec:probabilistic-model}
Our approach uses retrieval as a way to augment input examples at the granularity of small chunks of tokens.
Formally, we consider sequences of integer tokens in $\VV = [1, v]$, obtained using a text tokenizer\footnote{We use the notation $[1, v] \triangleq \{1, \dots, v \}$ throughout the text.}.
We split each $n$-token-long example $X = (x_1, \dots, x_n)$ into a sequence of $l$ chunks $(C_1, \dots, C_l)$  of size $m=\frac{n}{l}$, i.e. $C_1  \triangleq {(x_1, \dots, x_m),\, \dots ,\, C_{l} \triangleq} (x_{n - m + 1}, \dots, x_n)  \in \VV^{m}$. We use $n=2048$ and $m=64$.  
We augment each chunk $C_u$ with a set $\ret_\Dd(C_u)$ of $k$ neighbours from the database $\Dd$. 
$\ret_\Dd$ (or $\ret$ for brevity) is a non-trainable operator specified in \autoref{subsec-data-preparation}. 
Token likelihoods are provided by a model, parameterized by $\theta$, that takes as input both previous tokens and their retrieved neighbours. This defines the following retrieval-enhanced sequence log-likelihood:

\begin{equation}\label{eq:likelihood}
    L\left(X |\theta, \Dd \right)
    \triangleq \sum_{u=1}^{l} \sum_{i=1}^{m} \ell_\theta \left (x_{(u-1)\,m + i} | (x_j)_{j < (u-1)\,m + i},\ (\ret_\Dd(C_{u'}))_{u' < u} \right).
\end{equation}

We set $\ret(C_1) = \emptyset$, namely the likelihood of tokens from the first chunk does not depend on any retrieval data.
This likelihood definition preserves \textit{autoregressivity}: the probability of the $i$-th token of the $u$-th chunk, $x_{(u-1) m + i}$, only depends on previously seen tokens $(x_j)_{1 \leq j < (u-1)m + i}$ and on the data retrieved from the previous chunks
$(\ret(C_{u'}))_{u'<u}$. 
We can therefore directly \textit{sample} with log-probability $\ell$, where sampling within the chunk $C_u$ is conditioned on the neighbours $(\ret(C_{u'}))_{u'<u}$. 
This makes retrieval-enhanced models directly comparable with the largest language models that are evaluated by sampling.

\subsection{Nearest neighbour retrieval}
\label{subsec-data-preparation}

\paragraph{Retrieval neighbours.} Our database consists of a key-value memory. Each value consists of two contiguous chunks of tokens which we denote $[N, F]$ where $N$ is the \textit{neighbour} chunk which is used to compute the key, and $F$ is its \textit{continuation} in the original document. The corresponding key is the \bert embedding of $N$, averaged over time, that we denote $\bert(N)$. For each chunk $C$, we retrieve its approximate $k$-nearest neighbours from our key-value database using the $L_2$ distance on BERT embeddings $d(C, N) = ||\bert(C) - \bert(N)||_2^2$. The model receives the corresponding values $\ret(C) \triangleq ([N^1, F^1], \dots,  [N^k, F^k])$. Both neighbour chunks and their continuations provide meaningful improvements, as illustrated in our ablation study (\autoref{app:ablation}). We use a length $64$ for both $N^j$ and $F^j$, thus $\ret(C)$ has a shape of $k \times r$ with $r=128$.
To avoid retrieving the chunk $C_{u+1}$ in the retrieval set $\ret(C_u)$, which would break causality during training, we filter out neighbours originating from the same document as the training sequence~$X$.  

For a database of $T$ elements, we can query the approximate nearest neighbours in $\mathcal{O}(\log T)$ time. We use the SCaNN library \citep{avq_2020} to achieve this. This means that we can query our $2$ trillion token database in $10\,\textrm{ms}$ whilst evaluating or sampling from the model; this expense is amortized over a chunk length.
Performing retrieval on-the-fly is too slow to keep up with the training calculations---we leverage the frozen aspect of the embedding operator $\bert$ to precompute all approximate nearest neighbours and save the results as part of the data. 
In \autoref{fig:supp_wiki} in the Appendix, we show results where we only retrieve neighbours within Wikipedia. We find that neighbours tend to come from 2-3 links away from a given article whereas random articles are more than 5 links apart.

\subsection{$\retro$ model architecture}
\label{subsec-model}
Our model relies on an encoder-decoder transformer architecture, integrating the retrieved data through a cross-attention mechanism as introduced in \citet{vaswani2017attention}.
First, the retrieved tokens $\ret(C)$ are fed into an encoder Transformer, which computes the encoded neighbours set $E$.
Denoting the intermediate activations by $H$, our transformer decoder then interleaves $\retro$-blocks $\retro (H, E)$ and standard Transformer blocks $\textsc{LM}(H)$ (the hyperparameter $P \subseteq [1, L]$ determines at which layers we use a $\retro$-block).
These blocks are built from three different residual operators with signature $\RR^{n \times d} \to \RR^{n \times d}$: a fully-connected layer $\ffw$, the standard sequence-level self-attention layer $\attn$, and a chunked cross-attention layer $\ccca( \cdot, E)$ that incorporates information from the retrieval encoder:
\begin{equation}
   \retro \left(H, E\right) \triangleq \ffw \left( \ccca\left( \attn\left(H\right) , E\right) \right),\quad\text{and}\quad
    \textsc{Lm}(H) \triangleq \ffw(\attn(H))
\end{equation}

Since $\ffw$, $\attn$ and $\ccca$ are all autoregressive operators whose output at position $i$ only depends on $(h_{j})_{j \leq i}$, any succession of $\retro$ and $\textsc{lm}$ layers, followed by a token classification head defines an autoregressive log-likelihood \eqref{eq:likelihood}. 
An overview of the model architecture is given in \autoref{alg:encoderdecoder} and in \autoref{fig:architecture-summary}.
We next describe the retrieval encoder and the chunked cross-attention layer in more detail, and explain how to sample from $\retro$.

\begin{algorithm}[h]
 \caption{Overview of $\retro$ model architecture.}\label{alg:encoderdecoder}
 \SetKwFunction{encoderfn}{$\encoder$}

  \SetAlgoLined
  \SetKwProg{Def}{def}{:}{}

  \DontPrintSemicolon
\SetNoFillComment
 \textbf{Hyperparam:} $P$ and $P_{\textrm{enc}}$, indices of layers with cross-attention in the decoder and encoder respectively \\
 \textbf{Hyperparam:} $L$ and $L_{\textrm{enc}}$, number of decoder layers and number of encoder layers.
 \\
 \KwIn{$X \in \VV^n$: sequence of tokens. ${(\ret(C_u))}_{1 \leq u \leq l}$: the retrieved neighbours}
 \KwOut{$O \in \RR^{n \times | \VV |}$: the output logits \newline}
 \Def{$\encoder(\ret(C_u)_{1 \leq u \leq l}, H)$}{

    ${(H_u)}_{u \in [1, l]} \gets \spli(H)$ \;
    \For(\tcp*[h]{Encoder shared across neighbours and chunks}){$j \in [1,k], u \in [1, l]$}{
      $E_u^j = \emb_{\textrm{enc}}(\ret(C_u)^j)$ \tcp*[l]{May be shared with the decoder $\emb$}
      \For{$p' \in [1, L_{\textrm{enc}}]$}{
        $E_u^j \gets \attn_{\textrm{enc}}(E_u^j)$ \tcp*[l]{Bi-directional attention}
        \If{$p' \in P_{\textrm{enc}}$}{
        $E_u^j \gets \crossattention_{\textrm{enc}}(E_u^j, H_u)$ \;
        }
        $E_u^j \gets \ffw_{\textrm{enc}}(E_u^j)$ \;
      }
    }
    \KwRet $E$\;
 }
 \;
 $H \gets \emb(X)$ \\
  \For{$p \in [1, L]$}{
        $H \gets \attn(H)$ \tcp*[l]{Causal attention}
        
        \If{$p = \min(P)$} { \tcp*[l]{The neighbour $\encoder$ is conditioned with the decoder activations of the last layer before the first cross-attention}
            $E$ = $\encoder(\ret(C_u)_{1 \leq u \leq l}, H)$
        }
        
        \If{$p \in P$}{
        $H \gets \ccca(H, E)$ \;
        }
        $H \gets \ffw(H)$ \;
      }
  $O \gets \readout(H)$ \;
\end{algorithm}

\paragraph{Encoding retrieval neighbours.} For each chunk $C_u$, the $k$ retrieval neighbours $\ret(C_u)$ are fed into a bi-directional transformer $\encoder$, yielding the outputs $E_u^j \triangleq \encoder(\ret(C_u)^j, H_u) \in \RR^{r \times d'}$, where $j \in [1, k]$ indexes each neighbour. The retrieval encoder is a non-causal transformer.
It is conditioned on $H_u$, the activations of chunk $C_u$, through cross-attention layers; this allows the representations of the retrieval encoder to be modulated by the retrieving chunk in a differentiable way. 
More precisely, the encoding of the $j\textsuperscript{th}$ neighbour of the $u\textsuperscript{th}$ chunk, $\ret(C_u)^j$, depends on the \textit{attended} activation $H_u \triangleq {(h_{(u-1) m + i})}_{i \in [1, m]}\in \RR^{m \times d}$ of chunk $C_u$ at layer $\min(P)$.
All neighbours for all chunks are encoded in parallel, yielding a full encoded set $E \triangleq {(E_u^j)}_{u \in [1, l], j \in [1, k]} \in \RR^{l \times k \times r \times d'}$.
We denote $E_u \in \RR^{k \times r \times d'}$ as the encoded neighbours for chunk $u \in [1, l]$.

\paragraph{Chunked cross-attention.}
To perform the $\ccca$ operation, we first split a given intermediate activation $H \in \RR^{n \times d}$ into $l{-}1$ \textit{attending chunks} ${\left(H_u^+ \triangleq {(h_{u\,m +i - 1})}_{i \in [1, m]} \in \RR^{m \times d} \right)}_{u \in [1, l-1]}$, as depicted on the right of \autoref{fig:architecture-summary}.
$H_u^+$ holds the intermediary embeddings of the last token in chunk $C_u$ and of the first $m-1$ tokens in $C_{u+1}$
\footnote{The last token of chunk $C_u$ is the first to be able to access the retrieved content $E_u$ while maintaining autoregressivity in \eqref{eq:likelihood}. Hence, there is a one token overlap between chunk $C_u =  { \left( x_{(u-1) m + i} \right) }_{i \in [1, m]}$ and the corresponding attending chunk $C_u^+ \triangleq { \left (x_{u \, m + i -1} \right )}_{i \in [1, m]}$.
}. 
We compute the cross-attention between $H_u^+$ and $E_u$---the encoded retrieval set obtained from chunk $C_u$. Attention is computed across time and across neighbours simultaneously, as we merge the neighbour and time dimensions of $E_u$ before applying cross-attention.
Since there is a notion of alignment between data chunks and retrieval neighbours, we use relative positional encodings as described in \autoref{app:relative_pos}.

We concatenate the $l{-}1$ outputs of the per-chunk cross-attentions (each of shape $m \times d$) across time, and properly pad the result; we thus form the output activation $\ccca(H, E) \in \RR^{n \times d}$. Formally, for each chunk $C_u$ and for each token $i \in [1, m]$ we set
\begin{equation}
    \ccca(H, E)_{u\, m + i - 1} \triangleq \crossattention(h_{u \,m + i - 1}, E_{u}),
\end{equation}
where $\crossattention$ is the cross-attention residual operator over time-concatenated encoded neighbours. We recall that this operator is defined in its simplest version by three parameter matrices $K \in \RR^{d \times c}, Q \in \RR^{d \times c}$ and $V \in \RR^{d \times d}$. For all $h \in \RR^d$ and $Y \in \RR^{T \times d}$, we define
\begin{equation}
    \crossattention(h, Y) \triangleq \softmax(Y K Q^T h) Y V,
\end{equation}
where the softmax is performed on the second dimension and all products are matrix products. We use multi-head cross-attention, and add positional encodings to the \softmax (see \autoref{app:relative_pos}).

The first $m-1$ tokens cannot attend to any neighbour of a previous chunk; at these positions, we define $\ccca$ as the identity, setting ${\ccca(H, E)}_j \triangleq h_j$ for all tokens $j \in [1, m-1]$. 
Finally, the last token $h_{l m}$ attends to the last retrieval set $E_l$ and we set $h_{l\,m} \triangleq \crossattention(h_{l\,m}, E_l)$ (not shown in \autoref{fig:architecture-summary}). 
\autoref{alg:pseudo_cca} contains a simplified implementation of $\ccca$.
Note that chunked cross-attention is autoregressive: the output of $\ccca$ at position $i$ depends on the sequence from tokens from $0$ to $i$ that is input to $\ccca$.

With $\retro$ models, even though each $\ccca$ cross-attention attends only to the neighbours of the preceding chunk $\ret(C_{u-1})$, the dependencies over previous neighbours are propagated via the self-attention operations. 
The activations of the $i$\textsuperscript{th} token in the $u$\textsuperscript{th} chunk therefore potentially depend upon the set of \emph{all} previous neighbours $\ret(C_{u'})_{u'<u}$, without incurring the quadratic cost of cross attending to that set.   

\paragraph{Sampling.} When sampling, at the end of a chunk $C_u$, we use SCaNN to retrieve neighbours $\ret(C_u)$, based on the embedding $\bert(C_u)$. The encoded neighbours $E_u = \encoder(\ret(C_u))$ are then used to condition the generation of the next chunk~$C_{u+1}$, which we do incrementally: overall the cost of sampling is thus quadratic in the size of the sampled sequence, as when sampling from regular Transformers; the added cost of retrieval is linear in the number of chunks $l$, and is negligible compared to the token sampling cost in practice.

\subsection{Baseline Transformer architecture}
We use a transformer \citep{vaswani2017attention} similar to the one described in \citep{radford2019language}, with some minimal changes: we replace LayerNorm with RMSNorm \citep{zhang2019root} and use relative position encodings \citep{dai2019transformer}. 
As baselines, we train retrieval-free transformers with 132M, 368M, 1.3B and 7.0B parameters (embedding matrices are excluded from parameter counts). The hyperparameters we used are detailed in \autoref{table-baseline-arch}. 
All retrieval models use the same size encoder for the retrieval data, with $d'=896$ and 2 layers, which roughly adds $19M$ parameters. The encoder uses relative positional encodings.
The retrieval models contain one $\retro$-block every 3 blocks, starting from layer 6.
For our smallest model, $\ccca$  is applied in layers 6, 9 and 12 of the main pathway and also once for query conditioning in the encoder, which adds an additional  $12M$ parameters. 
The relative number of extra parameters reduces as we increase the baseline model size.
All models are implemented using JAX \citep{jax2018github} and Haiku \citep{haiku2020github}. 

\begin{table}[b]
    \centering
    \caption{\textbf{Number of parameters} for our baseline and $\retro$ models, excluding embeddings, along with the corresponding hyperparameters.}
    \vspace{2.0mm}
    \begin{tabular}{r r r r c c c}
        Baseline parameters & $\retro$ & $\dmodel$ & $\dffw$ & \# heads & Head size & \# layers \\ 
        \toprule
        132M & 172M (+30\%) & 896 & 3,584 & 16 & 64 & 12 \\
        368M & 425M (+15\%) & 1,536 & 6,144 & 12 & 128 & 12 \\
        1,309M & 1,451M (+11\%) & 2,048 & 8,192 & 16 & 128 & 24 \\
        6,982M & 7,532M $\ $ (+8\%)  & 4,096 & 16,384 & 32 & 128 & 32\\
    \end{tabular}
\label{table-baseline-arch}
\end{table}

\subsection{Quantifying dataset leakage exploitation}\label{sec:leakage-quantification}

\retro models may arguably benefit more easily from evaluation dataset leakage, i.e. the fact that we evaluate on data that were also present in the training set. To better understand how retrieval improves language modelling performance, we therefore quantify evaluation likelihood as a function of the overlap between the evaluation and training datasets.

The following approach can be used with any language model, and depends only on the frozen retriever system presented in \autoref{subsec-data-preparation}. We split the evaluation sequences ${(X_i)}_i$ into chunks of length $m \le 64$, and we see the training data as a set of chunks $\Cc$.
For each evaluation chunk $C \in \Cc$, we retrieve the 10 closest neighbours (of length up to 128) in the training data.
We then compute the longest token substring common to both the evaluation chunk and its neighbours. 
This gives a number $s \in [0, m]$. 
The value $r(C) = \frac{s}{m}$, ranging from $0$ (chunk never seen) to $1$ (chunk entirely seen), gives a reliable indication of how much overlap there is between the evaluation chunk and the training data. 
For a given model, we then obtain the log-likelihood $\ell(C)$ of each chunk $C$, and the number of bytes $N(C)$ it encodes.
We then consider the filtered bits-per-bytes of the model:
\begin{equation}
    \forall\,\alpha \in [0,1],\quad \mathcal{C}_\alpha \triangleq \{ C \in \mathcal{C}, r(C) \leq \alpha \},\quad\textrm{bpb}(\alpha) \triangleq
    \frac{\sum_{C \in \mathcal{C}_\alpha} \ell(C)}{
    \sum_{C \in \mathcal{C}_\alpha} N(C)
    },
\end{equation}
which correspond to the bits-per-bytes on the set of chunks that overlap less than $\alpha\:\%$ with the training chunks. 
Note that the full evaluation bit-per-bytes performance is recovered by $\mathrm{bpb}(1)$. The function $\bpb(\cdot)$ allows us to evaluate the impact of evaluation leakage over predictive performance: for low $\alpha$, $\bpb(\alpha)$ gives an indication on how the model performs on chunks that are entirely new; the slope of $\bpb(\cdot)$ shows how much the model exploits evaluation leakage.

\section{Related Work}
\label{sec-related}

We first review existing work on using retrieval for language modelling, and compare \retro to these works (see \autoref{tab:method-comparison}).
As we train \retro models on a large dataset containing a substantial section of the internet, our work raises potential privacy, safety, and fairness issues that we then review. 

\begin{table}[b]
\caption{\textbf{Comparison of \retro with existing retrieval approaches.}}
\small
\vspace{2.0mm}

\label{tab:method-comparison}

\resizebox{\linewidth}{!}{
\begin{tabular}{  l  c c c c } 

 & \# Retrieval tokens & Granularity & Retriever training & Retrieval integration \\
\toprule

Continuous Cache & $\Oo \left(10^3 \right)$ & Token & Frozen ($\lstm$) & Add to probs \\ 

$\knnlm$ & $\Oo \left(10^9 \right)$ & Token & Frozen (Transformer) & Add to probs \\

$\spalm$ & $\Oo \left(10^9 \right)$ & Token & Frozen (Transformer) & Gated logits \\

$\dpr$ & $\Oo \left(10^9 \right)$ & Prompt & Contrastive proxy & Extractive QA \\ 

$\realm$ & $\Oo \left(10^9 \right)$ & Prompt & End-to-End & Prepend to prompt \\ 

$\rag$ & $\Oo \left(10^9 \right)$ & Prompt & Fine-tuned $\dpr$ & Cross-attention  \\ 

$\fid$ & $\Oo \left(10^9 \right)$ & Prompt & Frozen $\dpr$ & Cross-attention  \\ 

$\emdr$ & $\Oo \left(10^9 \right)$ & Prompt & End-to-End (EM) & Cross-attention  \\ 

\textbf{\retro (ours)} & $\mathbf{\Oo \left ({10^{12}} \right)}$ & \textbf{Chunk} & \textbf{Frozen ($\bert$)} & \textbf{Chunked cross-attention}   \\ 

\end{tabular}
}
\end{table}

\subsection{Retrieval for language modelling}

\citet{brants2007_large_language_models_in_mt} show that scaling the training data to trillions of tokens improves the machine translation performance of $n$-gram models. 
More recently, GPT-2 \citep{radford2019language}, GPT-3 \citep{brown2020language}, and Jurassic-1 \citep{jurassic} show that scaling up language models leads to massive improvements on many downstream tasks. 
At the same time, \citet{carlini2021extracting} demonstrate that large-scale language models can perfectly memorise parts of their training data, suggesting that enhancing models with retrieval may lead to further improvements.
However, significant leakage between train and test datasets \citep{lee2021deduplicating, lewis2020question} makes comparing and evaluating large models trained on large datasets difficult, especially once retrieval capabilities over the training dataset are added. 

Historically, information retrieval for text relies on inverted index matching such as TF-IDF and BM25 \citep{robertson2009probabilistic}. Foundational work use latent topic modelling approaches like LDA \citep{blei_latent_2003} to identify relevant neighbours \citep{wei_lda-based_2006}.
Work in machine translation such as \citet{zhang2018guiding} and \citet{gu2018search} retrieve translation pairs based on edit distance between source sentences and guide the translation output using the closest retrieved target sentences.
The retrieval database may also be structured --- for example, \citet{ahn2016neural} use a symbolic knowledge graph to improve an RNN language model.

With the success of deep learning, retrieving systems have partly switched to dense learned representations based on a neural network's activations.
Continuous cache \citep{grave2016improving} adds probability mass to tokens for which previous activations resemble the current activation vector, extending the model's context to the local history.
$\knnlm$ \citep{khandelwal2020generalization} applies this idea to transformers and extends the retrieval database to English Wikipedia, resulting in substantial improvements on Wikitext103 evaluation.
Continuous cache and $\knnlm$ do not modify the underlying neural-network models, but interpolate at inference between the language model's output and distributions computed from retrieved tokens.
These methods can therefore be plugged into any model without additional training, although this limits the model's ability to reason about the retrieved text.
$\spalm$ \citep{yogatama2021adaptive} addresses this limitation by adding an extra gating network to post-process the retrieved data; yet most of the network is unaffected by the retrieval during inference.

The retrieval representations may be trained directly instead of relying on a pre-trained model---retriever systems have been developed for this purpose, primarily on open-domain question answering. 
For example, $\dpr$ \citep{karpukhin2020dense} trains two $\bert$ models (for queries and keys respectively) using a contrastive loss to align the representations of a question and of its answers. \cite{lee_latent_2019} use an inverse cloze task to find semantic representations of passages for retrieval.
These works differs from continuous cache and $\knnlm$ in that they embeds passages (or chunks) of text together, as opposed to each token individually. The retriever network is trained in isolation of the downstream task that uses the retrieval data. This potential issue is specifically addressed by $\realm$ \citep{guu2020retrieval}, which trains the retrieval system end-to-end to maximize the final training cross-entropy.
This comes with the extra complexity of searching the database during training and periodically updating the embedding table, severely limiting the scale at which it can operate.
$\rag$ \citep{lewis2020retrieval} and $\fid$ \citep{izacard2020leveraging} build upon $\dpr$ to set the state of the art on question answering benchmarks by training encoder-decoder transformer models.
More recently, $\emdr$ \citep{emdr} extends $\fid$ by using an expectation-maximization algorithm to train the retriever end-to-end and achieves state of the art results compared to similarly sized models.

In the open-domain dialogue setting, BlenderBot 2.0 \citep{komeili2021internet} learns to issue textual internet queries, outperforming dense retrieval methods when evaluated on a task measuring how close model responses are to those of humans. This involves collecting a dataset of human dialogues with associated search queries, which limits the scalability of this approach.
\citet{hashemi2020guided} introduce the Guided Transformer, a modified Transformer similar to \retro, for document retrieval and clarifying question selection.
Although effective on question answering and other tasks with strong conditioning, none of these methods are designed to model arbitrary text sequences, in contrast with \retro.

$\retro$ shares components with $\knnlm$ and $\dpr$ in that it uses frozen retrieval representations.
\retro models longer sequences than QA examples; this requires to reason at a sub-sequence level, and to retrieve different documents for the different chunks of a sequence.
Similar to $\fid$, $\retro$ processes the retrieved neighbours separately in the encoder, and assemble them in the chunked cross-attention.
This differs from e.g. \realm, that prepends retrieved documents to the prompt.
Using chunks allows for repeated retrieval whilst generating a sequence as opposed to retrieving only once based on the prompt alone.
Furthermore, retrieval is done during the whole pre-training process in \retro, and is not simply plugged-in to solve a certain downstream task.
Finally, previous methods based on dense query vectors use small models and retrieval datasets with less than 3B tokens (English Wikipedia). \autoref{tab:method-comparison} summarizes the difference of \retro with existing approaches.

\subsection{Privacy, safety and fairness}

\citet{bender2021dangers, weidinger2021harms} highlight several dangers of large language models. Those stem from their ability to memorise training data, their high training cost, the static nature of their training data \citep{lazaridou2021pitfalls}, their tendency of amplifying inherent biases in the training data, and their ability to generate toxic language \citep{gehman2020realtoxicityprompts}. 
In this section we inspect these dangers, focusing on how retrieval augmented language models may exacerbate or mitigate them.

Large language models can perfectly memorise parts of their training data \citep{carlini2021extracting}.
When coupled with large training datasets gathered from the web or other sources, this has clear privacy and safety implications.
Retrieval models such as $\retro$ that have access to the entire training dataset during inference exacerbate these privacy issues by being able to directly copy training data.
However, retrieval systems offer a path towards mitigating these concerns via obliteration of the retrievable data at inference time.
In addition, differential privacy training \citep{abadi2016differentialprivacy} of retrieval models could guarantee that no private information is stored in the model weights, while individualisation on private data could be made by updating the retrieval database at inference time.

Due to their high training cost, re-training large language model regularly to incorporate new data, languages, and norms is prohibitively expensive.
To keep retrieval models up-to-date, it may be sufficient to update the retrieval database, which is orders of magnitude cheaper than re-training a model from scratch.
In addition to the benefits of updating models in terms of fairness and bias, simply training large language models has a significant energy cost \citep{strubell2019energy, green_ai}.
Retrieval mechanisms offer a path to reducing the compute requirements needed to train and update language models that reach a certain performance.

Large language models are prone to generating toxic outputs, as shown in  \citet{gehman2020realtoxicityprompts}.
\citet{bender2021dangers, jo2020lessons} advocate for the importance of better training data curation and documentation.
Additionally, if portions of the training data are found to be eliciting biased or toxic outputs after training, retrieval allows for some correction, as the offending retrieval data can be retroactively filtered.
However, it is also the case that without careful analysis and intervention, retrieval models may exacerbate biases that are present in the training data.
Retrieval models can also add a further source of bias through the selection mechanism for retrieval documents.
Further work in this area is required to better understand how retrieval affects the bias and toxicity of the model outputs.

Finally, samples from large models are difficult to interpret, making mitigating these issues all the more challenging \citep{belinkov-etal-2020-interpretability, jain-wallace-2019-attention}.
Retrieval provides more insights in to the outputs of a model, as one can directly visualise or modify the neighbours that are being used. The examples in \autoref{tab:beaver}, \ref{tab:sample_hamlet}, \ref{tab:sample_human_right} and \ref{tab:sample_pi} illustrate how retrieval makes language models more factual and interpretable by providing more transparent outputs.

\section{Results}

We first report results on language modelling benchmarks. Second, we show how to \retrofit pre-trained Transformer language models into retrieval models with few additional FLOPs. Next, we report $\retro$ results on question answering. Finally, we report evaluation metrics with leakage filtering, to better understand the source of the gains with retrieval.

\label{sec-results}
\subsection{Language modelling}
\label{subsec-scaling}

\paragraph{Datasets.}
We evaluate our models on C4 \citep{raffel2019t5}, Wikitext103 \citep{merity2016pointer}, Curation Corpus \citep{curationcorpusbase2020}, Lambada \citep{paperno2016lambada} and the Pile \citep{pile}. 
We also evaluate on a set of manually selected Wikipedia articles that were added or heavily edited in September 2021, months after our pre-training and retrieval dataset was collected (details are given in \autoref{appendix:wikipedia-sept21-desc}).
We construct the dataset with articles from the ``future'' and manually remove new articles that strongly overlap documents in our training data. This guarantees that the evaluation documents are not leaked in our training data.

For C4, Wikitext103, the Pile, and our Wikipedia dataset we evaluate the language modelling performance on entire documents and measure the bits-per-byte (bpb). 
We favour bits-per-byte over loss as it is tokenizer agnostic.
We evaluate with a sequence length of 2048 tokens but use a stride of 1024 within documents to mitigate boundary effects.
On Curation Corpus we concatenate the article, the ``\texttt{TL;DR:}'' string, and the summary, but only evaluate the bpb on the summary.
For Lambada we evaluate the accuracy on the last word, using greedy generation.

\begin{figure}
    \centering
    \includegraphics[width=.99\textwidth]{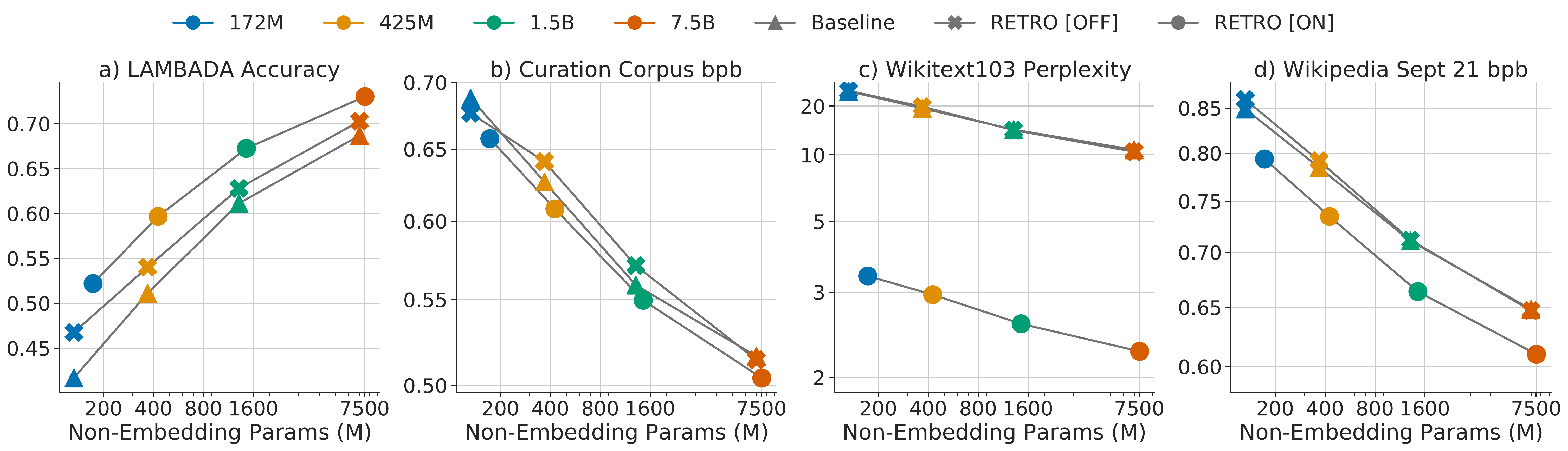}
    \caption{\textbf{Scaling with respect to model size.} (a) LAMBADA top-1 accuracy.  (b) Evaluation loss on curation corpus. (c) Perplexity on Wikitext103 valid. (d) Bits-per-byte on selected Wikipedia articles from September 2021.}
    \label{fig:scaling-wrt-params}
\end{figure}

\paragraph{Model scaling.} In \autoref{fig:summary}(left) and \autoref{fig:scaling-wrt-params} we show the language modelling performance as we scale models from 150 million to 7 billion (non-embedding) parameters.
We see that on all datasets, $\retro$ outperforms the baseline at all model sizes.
Furthermore, we observe that improvements do not diminish as we scale the models.
The performance is dataset dependent, with the largest gains on Wikitext103 and C4.
Wikipedia articles and other web pages are similar to Wikitext103 documents, even if not exact copies (\autoref{subsec-analysis-train-valid-leak}), we thus obtain dramatic improvements on Wikitext103 as our retrieval model is able to directly exploit these overlaps.
The smallest gains are for Curation Corpus, where $\retro$ only slightly outperforms the baseline. 
This is expected as Curation Corpus summaries are designed to only contain information from the source article and are not included in our retrieval database.
On our ``future'' Wikipedia September 2021 dataset, we also observe consistent gains for all model sizes.

\paragraph{Data scaling.} \autoref{fig:summary} (middle) shows how scaling the retrieval database at evaluation improves the language modelling performance.
We observe dramatic gains as the retrieval data is increased from Wikipedia (4 billion tokens) to all of Massive text (1.7T tokens).
\autoref{fig:summary}(right) shows how performance scales as we increase the number of retrieved chunks. 
Despite being only trained with 2 neighbours, we see consistent improvements for all models when the number of neighbours is increased from 1 to 10. 
Furthermore, we observe that larger models are able to better utilise more neighbours: the 172M model improves with up to 10 neighbours, whereas the 7B model improves with up to 40 neighbours.

\begin{figure}
    \centering
    \includegraphics[width=1.0  \textwidth]{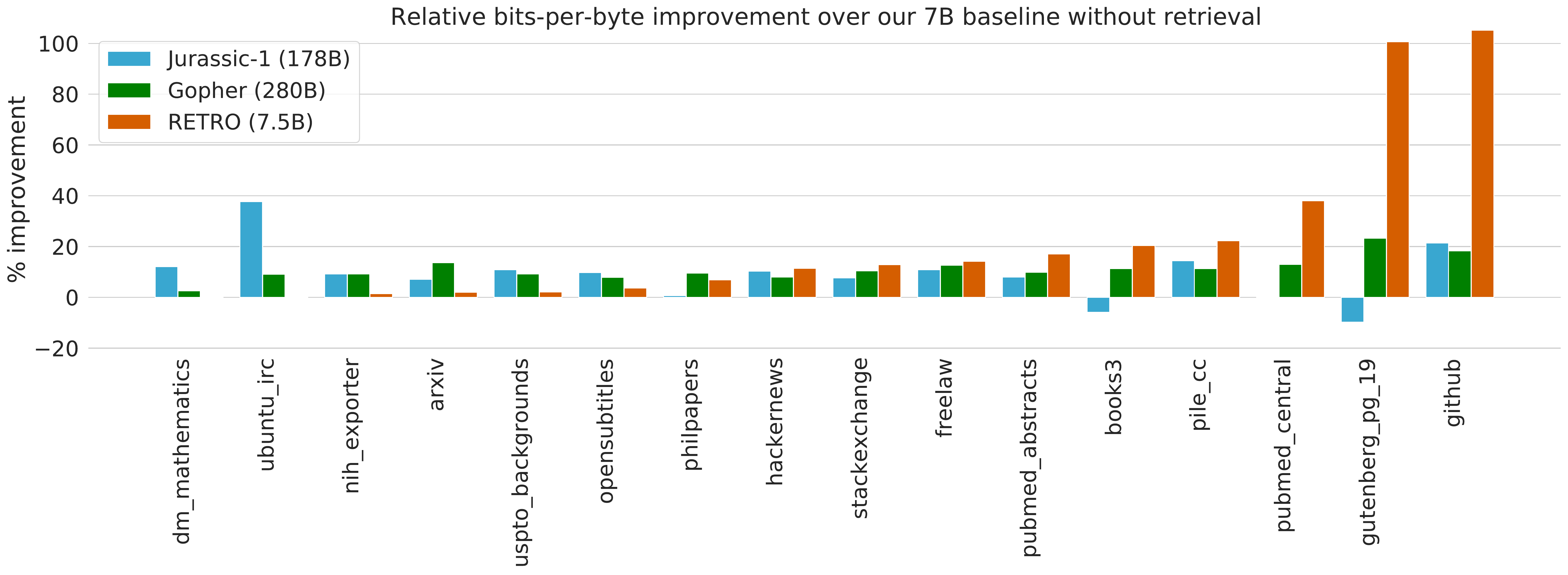}
    \caption{\textbf{The Pile: Comparison of our 7B baseline against Jurassic-1, Gopher, and $\retro$.} We observe that the retrieval model outperforms the baseline on all test sets and outperforms Jurassic-1 on a majority of them, despite being over an order of magnitude smaller.}
    \label{fig:pile_results}
\end{figure}

\paragraph{The Pile.} We evaluate our 7B models on the Pile test sets\footnote{Due to legal and ethical concerns relating to their use, we exclude the Enron Emails and the Youtube Subtitles datasets.} and compare against the 178B parameter Jurrasic-1 \citep{jurassic} model and the 280B parameter Gopher \citep{rae2021gopher} model. We do not compare against GPT-3 as it is outperformed by Jurassic-1 and Gopher on almost all subsets.
\autoref{fig:pile_results} shows the relative improvements in bits-per-byte over our 7B transformer baseline for our 7.5B \retro model, Jurassic-1 and Gopher. 
Jurassic-1 outperforms the baseline on all datasets except for books, likely due to the inclusion of books in our training data. Gopher and \retro outperform the baseline on all test sets.
Overall, \retro 7.5B outperforms Jurassic-1 and Gopher on a majority of the test sets.
On the \texttt{dm\_mathematics} and \texttt{ubuntu\_irc} subsets, our \retro model does not outperform our 7B baseline and underperforms Jurassic-1. 
We hypothesise that the retrieved neighbours on these datasets are not helpful, due to a combination of what is in our retrieval dataset and the efficacy of the nearest-neighbour search.

\begin{table}[b]
\small{
    \begin{center}
    \caption{\textbf{Perplexities on Wikitext103.} When using the Wikpedia dataset for retrieval, $\retro$ performs similarly to our implementation of $\knnlm$. As we scale the retrieval dataset, $\retro$ performs much better. The perplexities for retrieving from full MassiveText are quite low, which is partly due to partial overlap with Wikitext103 not caught by our deduplication.}
    \vspace{2.0mm}
    \resizebox{\linewidth}{!}{
        \begin{tabular}{l l  r r r r r}
        Model  & Retrieval Set & \#Database tokens & \#Database keys & Valid & Test  \\
        \toprule
       Adaptive Inputs~\citep{baevski2019adaptive} &   - &    - & -      & 17.96 & 18.65  \\
       $\spalm$~\citep{yogatama2021adaptive} & Wikipedia & 3B & 3B & 17.20 & 17.60\\
       $\knnlm$~\citep{khandelwal2020generalization} & Wikipedia  & 3B & 3B & 16.06 & 16.12 \\
       Megatron~\citep{megatron} & - & -  & - & - & 10.81 \\ 
      
      \midrule
       Baseline transformer (ours) &  -                            & - & -  & 21.53 & 22.96 \\
       $\knnlm$ (ours) &  Wikipedia                                & 4B & 4B & 18.52 & 19.54\\
       $\retro$ & Wikipedia                                       & 4B & 0.06B & 18.46 & 18.97\\
       $\retro$ & C4                                                 & 174B & 2.9B & 12.87 & 10.23 \\
       $\retro$ & MassiveText (1\%)                                    & 18B & 0.8B & 18.92  & 20.33\\
       $\retro$ & MassiveText (10\%)                                   & 179B & 4B & 13.54 & 14.95\\
       $\retro$ & MassiveText (100\%)                & 1792B& 28B  & \textbf{3.21} & \textbf{3.92}\\

    \end{tabular}}
    \label{table-wikitext103}
    \end{center}
    }
\end{table}

\paragraph{Wikitext103.} To validate our approach in a controlled setting, we compare our method with $\knnlm$ \citep{khandelwal2020generalization} on the Wikitext103 dataset in \autoref{table-wikitext103}. 
We train a baseline transformer on the training set of Wikitext103. This transformer has 24 layers, 1024 hidden units, 16 heads and a key size of 64, as in \cite{baevski2019adaptive}. Our baseline does not have adaptive input, and our tokenizer has an open vocabulary, unlike \cite{baevski2019adaptive}, which makes our baseline perplexities a bit higher. The full experiment details and hyperparameters are given in \autoref{subsubsec-app-wikitext103} and \autoref{tab:app-wikitext103-hyper}.

We re-implement $\knnlm$ with our tokenizer and baseline transformer to produce embeddings of size 1024 for every token in Wikitext103. $\knnlm$ has probabilities $p_{\knnlm}=  \lambda p_{\knn} + (1-\lambda) p_{\lm}  $ with $p_{\knn} \left (n_k \right ) \propto \exp \left (-\alpha d_k \right )$. We tune $\lambda=0.118$ and $\alpha=0.00785$ on the validation set (\autoref{fig:app_wikitraining_curves}) and report performance for these hyperparameters on both the validation and test set.

We fine-tune our baseline transformer into a $\retro$ model (\autoref{fig:app_wikitraining_curves}), using the Wikitext103 training data and retrieving from Wikipedia with 2 neighbours. 
We only train the new weights, as explained in \autoref{subsec:ft}, and share the embedding weights between the encoder and the main pathway. This is necessary for Wikitext103 which is quite small, as training $\retro$ from scratch in this setting leads to over-fitting. 

We evaluate the fine-tuned $\retro$ model with different retrieval sets.
We use 10 neighbours at evaluation for both $\retro$ and $\knnlm$.  
When retrieving from Wikipedia, we obtain results comparable to our $\knnlm$ implementation. 
Furthermore, scaling the retrieval database to MassiveText yields dramatic improvements, though this is partly due to leakage (see \autoref{subsec-analysis-train-valid-leak}).
For reproducibility, we also include results when retrieving from C4, which are close to previous state-of-the-art and comparable to using 10 \% of MassiveText. 

It is worth noting that $\knnlm$ requires 1024 floats for every token in the retrieval dataset, totalling 15 terabytes (Tb) for the 4 billion tokens in Wikipedia.
$\knnlm$ and other token-level retrieval approaches therefore don't scale to retrieval databases with trillions of tokens such as MassiveText.
In comparison, $\retro$ only requires 215Gb to index our Wikipedia dataset, and 93Tb for MassiveText.
Inspecting the number of retrieval database entries in \autoref{table-wikitext103} makes it clear why retrieving at the chunk level is necessary when scaling to datasets with trillions of tokens.

\subsection{\retro-fitting baseline models}
\label{subsec:ft}
\begin{figure}
    \centering
    \includegraphics[width=1.0  \textwidth]{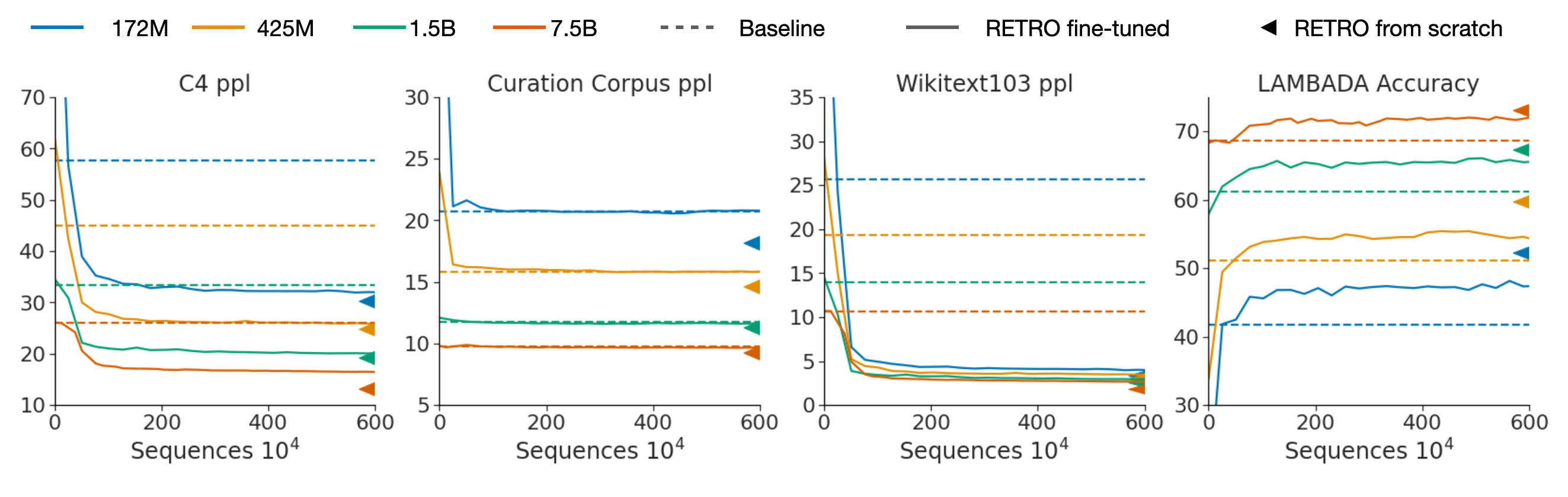}
   
    \caption{\textbf{$\retro$-fitting a baseline transformer.}
    Any transformer can be fine-tuned into a retrieval-enhanced transformer by randomly initializing and training only the chunked cross-attention and retrieval encoder weights. Fine-tuning in this way quickly recovers and surpasses the non-retrieval performance, and almost achieves the same performance as training a retrieval model from scratch (shown by the arrow on the right hand side of each plot).
    We find good performance $\retro$-fitting our models training on only 3\% the number of tokens seen during pre-training. }
    \label{fig:ft}
\end{figure}

We extend baseline models into $\retro$ models by freezing the pre-trained weights and training only chunked cross-attention and neighbour encoder parameters (less than 10\% of weights for the 7B model) in \autoref{fig:ft}. 
This offers an efficient alternative path to enhance transformers with retrieval, requiring only 6 million sequences (3\% of the pre-training sequences that we used).
Additionally, by only training the new weights we ensure that when evaluated without retrieval, the original model performance is exactly maintained.
\retrofitting models quickly surpasses the performance of baseline models and even achieves performance close to that of $\retro$ models trained from scratch.
The experiment hyperparameters are given in \autoref{appendix:retrofitting-details}.

\subsection{Question answering}
\label{subsection-qa}
We fine-tune our retrieval models on the Natural Questions \citep{naturalquestions} dataset to demonstrate that our retrieval pathway can be used to inject information from arbitrary data sources.
We use the version\footnote{\url{https://github.com/facebookresearch/FiD}} provided by \cite{izacard2020leveraging} which is augmented with the retrieved passages from $\dpr$ \citep{karpukhin2020dense}.
We fine-tune all the weights of our 7.5B pre-trained $\retro$ model for 25,000 steps using the top 20 retrieved passages. 
We format the data as  ``\texttt{question: \{question\} \textbackslash{}n\,answer: \{answer\}}'' and left pad the data such that ``\texttt{answer:}'' coincides with the end of the first chunk of 64 tokens and thus aligns with the first retrieving chunk. The model has access to the question via the previous tokens in the sequence as well as the top 20 DPR Wikipedia passages and their titles via the chunked cross-attention mechanism. The exact match scores are shown in \autoref{tab:natural-questions} and the full fine-tuning details are given in \autoref{appendix:natural-questions}.
Our method is competitive with previous approaches such as $\realm$, $\rag$ and $\dpr$, but underperforms the more recent $\fid$. In contrast with this work, we find that increasing the number of neighbours past 20 does not improve $\retro$ performance on this task. We hypothesise that the encoder-decoder structure of T5---the base model in $\fid$--- and the T5 pre-training objective leads to a model that relies more on the encoder output than $\retro$, which is important in the QA setting. To compete with T5-finetuned models, future work should consider ways of forcing \retro to rely further on the retrieval encoder output when producing tokens.

\begin{table}[b]
    \caption{\textbf{Question answering results.} Exact match accuracy on Natural Questions.}
    \small
    \begin{center}
    \begin{tabular}{l c}
    Model & Test Accuracy\\ [0.5ex] 
    \toprule
    $\realm$ \citep{guu2020retrieval}    & 40.4 \\
    $\dpr$ \citep{karpukhin2020dense}    & 41.5 \\
    $\rag$ \citep{lewis2020retrieval}    & 44.5 \\
    $\emdr$ \citep{emdr}                 & 52.5 \\
    $\fid$ \citep{izacard2020leveraging} & 51.4 \\
    $\fid$ + Distill. \citep{izacard2020memory} & \textbf{54.7} \\
    \midrule
     Baseline 7B (closed book)         & 30.4 \\
     $\retro$ 7.5B (DPR retrieval)       & 45.5 \\
    \end{tabular}
    \end{center}
    \label{tab:natural-questions}
\end{table}

\subsection{Relating retrieval performance to dataset leakage.}
\label{subsec-analysis-train-valid-leak}

\begin{figure}
    \centering
    \includegraphics[width=\linewidth]{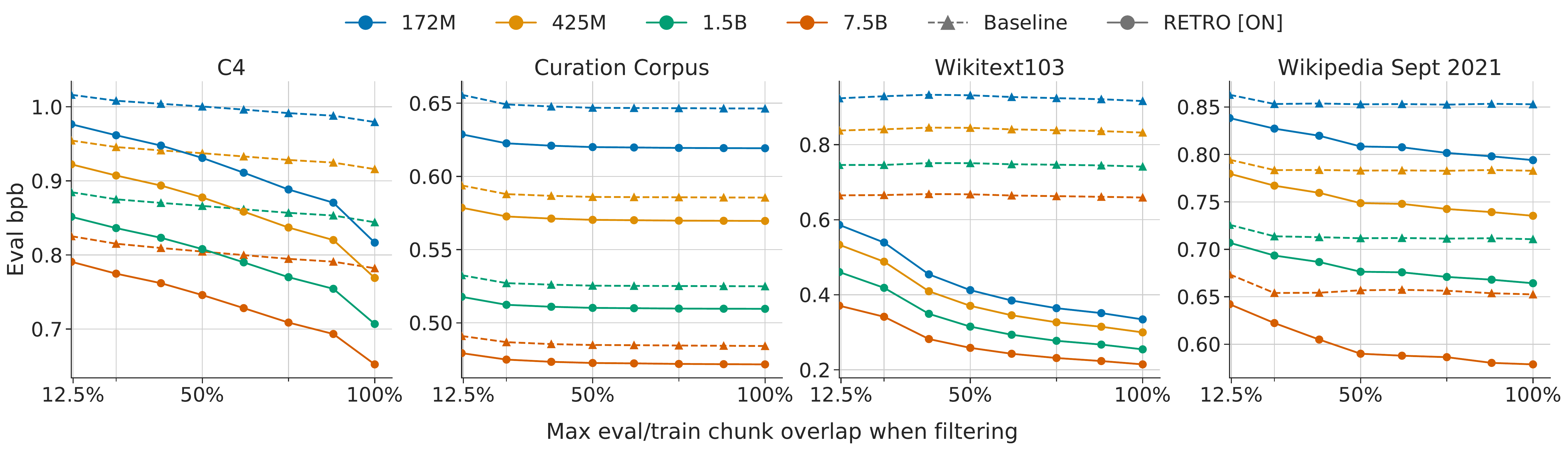}
    \caption{\textbf{Performance vs. longest common retrieval substring.} Evaluation loss as a function of allowed longest common substring between evaluation data chunks and their nearest neighbours. Retrieval still helps when considering chunks with no more than 8 contiguous tokens overlapping with training dataset chunks.}
    \label{fig:scaling-wrt-leakage}
\end{figure}

We report the filtered eval losses as detailed in \autoref{sec:leakage-quantification} on C4, Curation Corpus and Wikitext103 in \autoref{fig:scaling-wrt-leakage}. On C4 and Wikitext103, for which there is leakage into the training set, the slope is negative for both baseline models and \retro models. \retro models exploit leakage more strongly than baseline models, as indicated by the more negative slope. This is due to its explicit ability to copy-paste existing training chunks to predict leaked evaluation chunks (see a qualitative example of this model behavior on a Wikitext103 article in \autoref{tab:data_wiki103_radclif}). On Curation Corpus, retrieval provides a constant offset, which is expected as there is by design no leakage between Curation Corpus and the training dataset.

On the other hand, \retro outperforms baseline models at all leakage levels, down to $\alpha = 12.5\%$. At this level, the loss is computed on chunks with less than $8$ contiguous tokens shared with the closest matching chunk in the training dataset---this is a reasonable level of overlap at which we consider that there is no local leakage. Retrieval thus improves predictions on both chunks that are syntactically similar to chunks in the training set, and on chunks that are syntactically different from all training chunks. This points toward a non trivial \retro capacity of generalizing based on both model parameters and retrieval database. Similar results are found on the Pile dataset (see \autoref{fig:distribution_pile}, \autoref{app:leakage}).

\subsection{Using \retro for sampling}
We show examples of samples obtained using the 7.5B $\retro$ model in \autoref{tab:beaver}, \autoref{tab:sample_hamlet} and \autoref{app:qualitative}. For each chunk (the first one being the prompt), we juxtapose sampled chunks $C_u$ with retrieved neighbours $\ret(C_{u})$.
To give an indication of local overlap, we colour each sampled token in chunk $C_u$ based on the length of the longest common prefix (LCP) found in the retrieved chunks $\ret(C_{u-1})$.
Similarly, we colour the retrieved chunks based on the LCP in the sampled chunk.
For the sample in \autoref{tab:beaver}, for which we chose the prompt, we observe that the retrieved chunks influence the sample as there are overlaps between the sampled tokens and neighbour tokens.
Overall, retrieval reduces hallucinations (in line with the findings of \citet{shuster_retrieval_2021}) and makes the model more knowledgeable, when comparing with samples produced with retrieval disabled. In the sample in \autoref{tab:sample_hamlet}, the model recognises that the prompt is the beginning of the first scene of Hamlet and leverages retrieval data to continue it with only a few mistakes. We provide further examples in \autoref{app:qualitative}, including examples from the evaluation sets, as well as the detailed procedure used for colouring the tables.

\section{Conclusion}
We present Retrieval-Enhanced Transformers ($\retro$), a method for modelling arbitrary text sequences whilst retrieving from databases with trillions of tokens---scaling the data available to models by an order of magnitude compared to what is typically consumed during training.
$\retro$ models gains do not diminish for models with up to at least 7B parameters, and correspond to non-retrieval models with 10$\times$ more parameters on certain datasets. On Wikitext103 and the Pile, \retro outperforms previous models trained on large scale datasets. We also show that \retro is competitive on retrieval-intensive downstream tasks such as question answering. 

$\retro$ models are flexible and can be used without retrieval at evaluation and still achieve comparable performance to baseline models. Conversely, baseline models can be rapidly fine-tuned into $\retro$ models to obtain nearly the same performance as if trained from scratch.
Careful analysis shows that only a modest fraction of the gains obtained by $\retro$ are due to test set leakage. In general, we caution for such leakage in large-scale language datasets and suggest further work in better understanding the role of test set leakage in the performance of large-scale language models.

Overall, our work demonstrates at an unprecedented scale that semi-parametric approaches can provide an orthogonal, more efficient approach than raw parameter scaling as we seek to build more powerful language models.

\section*{Acknowledgements}
We would like to thank Nikolai Grigorev, Marc'aurelio Ranzato, Cyprien de Masson d'Autume, Po-Sen Huang, Johannes Welbl, Lisa Anne Hendricks, Ethan Perez, Jeff Stanway, Eric Noland, Gregory Wayne, John Jumper, Julian Schrittwieser, Lorrayne Bennett, Devang Agrawal, Dani Yogatama, Susannah Young, Nando de Freitas, Demis Hassabis, and Koray Kavukcuoglu for their help, advice and reviews. 
Additionally, we would like to thank Zonglin Li, David Simcha, and the ScaNN developers for their help.

 \begin{table}[h]     \tiny     \caption{\small \textbf{Sample - Beavers are interesting animals}. The $\retrooff$ sample  quickly diverges to other animals while the $\retroon$ sample tends to stay focused on the beaver topic due to neighbour conditioning.}     \vspace{2.0mm}     \centering     
\begin{tabular} { l l l l }

 \samplevisheader Beavers\ are\ interesting\ animals\ that & Beavers\ are\ interesting\ animals\ that & \colorbox{cm_12!40}{\strut .}\colorbox{cm_25!40}{\strut }\colorbox{cm_12!40}{\strut }\colorbox{cm_25!40}{\strut Be}\colorbox{cm_38!40}{\strut avers}\colorbox{cm_12!40}{\strut \ build}\colorbox{cm_25!40}{\strut \ their}\colorbox{cm_0!40}{\strut \ lodges}\colorbox{cm_12!40}{\strut \ in}\colorbox{cm_0!40}{\strut \ pon} & \colorbox{cm_0!40}{\strut naw\ them\ into\ smaller\ sections}\colorbox{cm_12!40}{\strut \ and}\colorbox{cm_0!40}{\strut \ d} \\ 
 \ live\ near\ rivers.\ They\ build & \ live\ near\ rivers.\ They\ build & \colorbox{cm_0!40}{\strut ds\ they\ have\ created}\colorbox{cm_12!40}{\strut \ in}\colorbox{cm_0!40}{\strut \ wooded\ areas} & \colorbox{cm_0!40}{\strut rag\ them\ into}\colorbox{cm_12!40}{\strut \ the}\colorbox{cm_0!40}{\strut \ water}\colorbox{cm_12!40}{\strut .}\colorbox{cm_25!40}{\strut }\colorbox{cm_0!40}{\strut }\colorbox{cm_12!40}{\strut }\colorbox{cm_0!40}{\strut Engineers} \\ 
  &  & \colorbox{cm_12!40}{\strut .}\colorbox{cm_25!40}{\strut }\colorbox{cm_12!40}{\strut }\colorbox{cm_0!40}{\strut Like\ many\ things}\colorbox{cm_12!40}{\strut \ in}\colorbox{cm_0!40}{\strut \ nature}\colorbox{cm_12!40}{\strut ,}\colorbox{cm_0!40}{\strut \ there} & \colorbox{cm_0!40}{\strut \ of}\colorbox{cm_12!40}{\strut \ the}\colorbox{cm_0!40}{\strut \ Pond}\colorbox{cm_12!40}{\strut \ Be}\colorbox{cm_25!40}{\strut avers}\colorbox{cm_0!40}{\strut \ are\ interesting} \\ 
  &  & \colorbox{cm_0!40}{\strut \ is\ a\ connection\ between\ creatures}\colorbox{cm_12!40}{\strut \ i} & \colorbox{cm_0!40}{\strut \ animals\ because\ they\ change}\colorbox{cm_12!40}{\strut \ the}\colorbox{cm_0!40}{\strut \ hab} \\
\end{tabular}
    \label{tab:beaver}
\end{table}
 \begin{table}[h]     \tiny     \tiny     \caption{\small \textbf{Sample - Hamlet, Act 1, Scene 1.}  The $\retrooff$ sample has correct syntax but is hallucinated, and ends with repetition of one character (\emph{FRANCISCO Approach me not}). The $\retroon$ sample is the correct continuation of the original text, and is robust to formatting differences between our prompt and the retrieved data.}     \vspace{2.0mm}     \centering     \begin{tabular}{ l  l  l  l} \samplevisheader ACT\ I\ SCENE\ I.\ Elsinore.\ A\ pla & ACT\ I\ SCENE\ I.\ Elsinore.\ A\ pla & \colorbox{cm_0!40}{\strut ACT}\colorbox{cm_12!40}{\strut \ I}\colorbox{cm_0!40}{\strut \ SCENE}\colorbox{cm_12!40}{\strut \ I.}\colorbox{cm_0!40}{\strut \ Elsinore}\colorbox{cm_12!40}{\strut .}\colorbox{cm_0!40}{\strut \ A\ pla} & \colorbox{cm_0!40}{\strut ><TEXT>ACT}\colorbox{cm_12!40}{\strut \ I}\colorbox{cm_0!40}{\strut \ \ SCENE}\colorbox{cm_12!40}{\strut \ I.}\colorbox{cm_0!40}{\strut \ Elsin} \\ 
 tform\ before\ the\ castle.\ FRANC & tform\ before\ the\ castle.\ FRANC & \colorbox{cm_0!40}{\strut tform\ before}\colorbox{cm_12!40}{\strut \ the}\colorbox{cm_0!40}{\strut \ castle}\colorbox{cm_12!40}{\strut .}\colorbox{cm_0!40}{\strut \ (FRAN} & \colorbox{cm_0!40}{\strut ore}\colorbox{cm_12!40}{\strut .}\colorbox{cm_0!40}{\strut \ A\ platform\ before}\colorbox{cm_12!40}{\strut \ the}\colorbox{cm_0!40}{\strut \ cas} \\ 
 ISCO\ at\ his\ post.\ Enter\ to\ him & ISCO\ at\ his\ post.\ Enter\ to\ him & \colorbox{cm_0!40}{\strut C}\colorbox{cm_12!40}{\strut ISCO\ at}\colorbox{cm_0!40}{\strut \ his\ post}\colorbox{cm_12!40}{\strut .}\colorbox{cm_0!40}{\strut \ Enter}\colorbox{cm_12!40}{\strut \ to}\colorbox{cm_0!40}{\strut \ hi} & \colorbox{cm_0!40}{\strut tle}\colorbox{cm_12!40}{\strut .}\colorbox{cm_0!40}{\strut \ }\colorbox{cm_12!40}{\strut \ FRANC}\colorbox{cm_25!40}{\strut ISCO}\colorbox{cm_12!40}{\strut \ at}\colorbox{cm_0!40}{\strut \ his\ post}\colorbox{cm_12!40}{\strut .}\colorbox{cm_0!40}{\strut \ E} \\
\end{tabular}
    \label{tab:sample_hamlet}
\end{table}

\bibliography{main}

\pagebreak
\appendix

\section{Datasets}

We provide a full description of MassiveText and of our extract of recent Wikipedia articles.

\subsection{Full description of MassiveText}
\label{appendix:massive-text}
The full break down of MassiveText by source and languages is given in \autoref{tab:massive-text-full}.
For a full description and analysis of MassiveText, see \citet{rae2021gopher}.

\begin{table}[h]
\small 
\begin{center}
\begin{tabular}{c c c r r r}
Source & Language & Token count (M) & Documents  & Sampling weight\\ [0.5ex] 
\toprule
\multirow{ 10}{*}{Web} & En &  483,002 &  604,938,816 & 0.314 \\ 
 & Ru &  103,954 &  93,004,882 & 0.033 \\ 
 & Es &  95,762 &  126,893,286 & 0.033 \\ 
 & Zh &  95,152 &  121,813,451 & 0.033 \\ 
 & Fr &  59,450 &  76,612,205 & 0.033 \\ 
 & De &  57,546 &  77,242,640 & 0.033 \\ 
 & Pt &  44,561 &  62,524,362 & 0.033 \\ 
 & It &  35,255 &  42,565,093 & 0.033 \\ 
 & Sw &  2,246 &  1,971,234 & 0.0044 \\ 
 & Ur &  631 &  455,429 & 0.0011 \\ 
\midrule
Books & En &  3,423,740 & 20,472,632  & 0.25 \\
\midrule
News & En & 236,918 & 397,852,713 & 0.1 \\
\midrule
\multirow{10}{*}{Wikipedia}  & En &  3,977 &  6,267,214 & 0.0285 \\ 
 & De &  2,155 &  3,307,818 & 0.003 \\ 
 & Fr &  1,783 &  2,310,040 & 0.003 \\ 
 & Ru &  1,411 &  2,767,039 & 0.003 \\ 
 & Es &  1,270 &  2,885,013 & 0.003 \\ 
 & It &  1,071 &  2,014,291 & 0.003 \\ 
 & Zh &  927 &  1,654,772 & 0.003 \\ 
 & Pt &  614 &  1,423,335 & 0.003 \\ 
 & Ur &  61 &  344,811 & 0.0001 \\ 
 & Sw &  15 &  58,090 & 0.0004 \\ 
\midrule
Github & - &374,952 & 142,881,832  & 0.05 \\
\midrule
Total & - & 5,026,463 & 1,792,260,998 & 1
\end{tabular}
\caption{\textbf{MassiveText dataset.} The final column indicates the sampling weight for each dataset during training. For the retrieval database, the entire dataset is used, with the exception of books for which we use a sub-sample of 4\%.}
\label{tab:massive-text-full}

\end{center}
\end{table}

\subsection{Wikipedia September 2021}
\label{appendix:wikipedia-sept21-desc}

\begin{table}[h]
\caption{Full set of articles included in our \textbf{Wikipedia Sept. 2021} evaluation dataset.}
\label{table:wikipedia-sept21-full}

\small

\begin{center}
\begin{tabular}{l l}
Megan Rohrer &
Aakashavaani \\
Emma Raducanu &
Junior Eurovision Song Contest 2021 \\
Ambra Sabatini &
Pavilion Bukit Jalil \\
WhyDonate &
Blake Desjarlais \\
The Juggernaut (company) &
2021 All-Ireland Senior Football Championship Final \\
Angela Diaz &
Drift-barrier hypothesis \\
2020 Summer Paralympics &
Venomics \\
2021 Afghan protests &
Great Circle (novel) \\
Rexh Xhakli &
Hurricane Ida \\
Julia Laskin &
2021 Montenegrin episcopal enthronement protests \\
Cuijk &
At War With the Silverfish \\
Ghoubet Wind Power Station \\
\bottomrule
\end{tabular}
\end{center}
\end{table}

We create an evaluation dataset consisting of 23 Wikipedia articles that were added or heavily edited in September 2021, after we collected our training dataset. In addition, we filter out articles that rely too heavily on templated content, using the method detailed in \autoref{sec:leakage-quantification} to identify articles with chunks that have a high overlap with their neighbours. \autoref{fig:leakage_distribution_main} show that little overlap remains between our test dataset and the retrieved neighbours from the training dataset.
The full list of included articles is given in \autoref{table:wikipedia-sept21-full}. 

We first parse articles using \texttt{mwparserfromhell}\footnote{\url{https://github.com/earwig/mwparserfromhell}}.
We then remove sections with the following titles: ``references'', ``external links'', ``sources'', ``further reading'', ``see also'', ``citations'', and ``note''.
In the remaining sections, we remove Wikilinks and remove the following templates: ``reflist'', ``notelist'', ``notelist-ua'', ``notelist-lr'', ``notelist-ur'', and ``notelist-lg''. We also exclude objects with the ``ref'' or ``table'' tag and clean the remaining text with the \texttt{strip\_code} function.
Finally, we concatenate the title and all the sections and use \texttt{\textbackslash{}n\textbackslash{}n} to delimitate them. 

\section{Details on the retrieval architecture}

We give details on the \retro architecture, and on the fine-tuning procedure we use for \retrofitting existing language models.

\subsection{\retro architecture and implementation}

\subsubsection{Feed-forward architecture}\label{app:feed_forward}

As mentioned in the main text, the overall encoder-decoder architecture is fully feed-forward. 
We start with a sequence $X \in \VV^n = {(C_u)}_{1 \leq u \leq l}$, and its pre-computed neighbours ${(\ret(C_u))}_{1 \leq u \leq l}$ and returns logits in $\RR^{n \times | \VV |}$. Along with $\attn$, $\ffw$, $\ccca$ and $\crossattention$ operators introduced in the main text, we define the decoder embedding layer $\emb: \VV^n \to \RR^{n \times d}$, the $\spli$ operator that extracts chunked intermediary embeddings $\spli(H) \triangleq {(H_u)}_{1 \leq u \leq l} \in \RR^{l \times m \times d}$ and the read-out layer $\readout: \RR^{n \times d} \to \RR^{n \times | \VV |}$. We then describe the forward pass in \autoref{alg:encoderdecoder}. In addition to the usual Transformer ones, $\retro$ architecture hyperparameters involves the layer indices $P_{\textrm{enc}}$ and $P$, at which the encoder and the decoder perform cross-attention.

\subsubsection{Relative positional encoding in the chunked cross-attention layer}\label{app:relative_pos}

The $\crossattention$ operator uses relative positional logits, that are computed from a specific relative distance separating data tokens from retrieval tokens. Indeed, we expect any retrieval neighbour $\ret(C_u)^j$ and the chunk $C_u$ to be relatively well aligned, and assume that they start at the same position. Therefore, when computing $\crossattention(H_u^+, E_u)$, we set the distance between the data token $i \in [1, l]$ of chunk $C_u^+$ and the retrieval token $i' \in [1, 2 l]$ of $\ret(C_u)^j$ to be 
\begin{equation}
    d(i, i') \triangleq i - i' + l - 1.
\end{equation}
When computing the encoder cross-attentions $\crossattention(\ret(C_u)^j, H_u)$, we set the distance between the retrieval token $i' \in [1, 2 l]$ and the data token $i \in [1, l]$ to be 
\begin{equation}
    d_{\textrm{enc}}(i', i) \triangleq i' - i.
\end{equation}
Positional logits are obtained as a linear transform of a cosine vector computed from $(d(i, i'))_{i, i'}$, and are added to content logits, as in a regular self-attention block.

\subsubsection{Chunked cross-attention implementation}\label{app:ccca}

Our implementation of the $\ccca$ operator, shown in \autoref{alg:pseudo_cca}, is based on a vectorized application of a cross-attention layer.
For simplicity, we omit the multi-head attention logic and use the simplest Q,K,V attention. 
We omit relative positional logits computation, described above.

\begin{lstlisting}[float,language=Python,
caption={Jax implementation of the \textbf{chunked cross attention}, simplified.\label{alg:pseudo_cca}},
basicstyle=\footnotesize\ttfamily,
morekeywords={self},              % Add keywords here
keywordstyle=\color{deepblue},
commentstyle=\color{blue},
emph={multi_chunk_cross_attention, multi_neighbour_cross_attention,cross_attention,relative_positional_encodings},          % Custom highlighting
emphstyle=\color{deepred},    % Custom highlighting style
stringstyle=\color{deepgreen},
frame=none,                         % Any extra options here
showstringspaces=false]

n = 128  # Sequence length
m = 16  # Chunk length
r = 32  # Retrieval length
k = 4  # Number of neighbours
d = 16  # Embedding size
l = n // m  # Number of chunks

# Parameters
Q = jnp.zeros((d, d))
K = jnp.zeros((d, d))
V = jnp.zeros((d, d))

def relative_positional_encodings(attending_length, attended_length):
  # Classical relative positional encodings
  ...

def cross_attention(chunk, neighbour):
  m, d = chunk.shape
  r, d = neighbour.shape
  queries = chunk @ Q
  keys = neighbour @ K
  logits = queries @ keys.T
  values = neighbour @ V
  return logits, values

def multi_neighbour_cross_attention(chunk, neighbours):
  m, d = chunk.shape
  k, r, d = neighbours.shape

  logits, values = jnp.vectorize(cross_attention, 
                                 signature='(m,d),(r,d)->(m,r),(r,d)')(
                                     chunk, neighbours)
  assert logits.shape == (k, m, r)
  assert values.shape == (k, r, d)
  logits += relative_positional_encodings(m, r)[None, :, :]
  logits = jnp.moveaxis(logits, 0, -1).reshape((m, r * k))
  values = jnp.moveaxis(values, 0, 1).reshape((r * k, d))
  return jax.nn.softmax(logits) @ values

def multi_chunk_cross_attention(observation, neighbours):
  attending_chunks = jnp.pad(observation[m-1:], 
                             ((0, m - 1), (0, 0)),
                             mode='constant').reshape(l, m, d)
  chunked_output = jnp.vectorize(multi_neighbour_cross_attention,
                                 signature='(m,d),(k,r,d)->(m,d)')(
                                     attending_chunks, neighbours)
  assert chunked_output.shape == (l, m, d)
  output = jnp.pad(chunked_output.reshape(n, d), 
                   ((m - 1, 0), (0, 0)),
                   mode='constant')[:n]
  return output


observation = jnp.zeros((n, d))  # Input
neighbours = jnp.zeros((l, k, r, d))

h = multi_chunk_cross_attention(observation, neighbours)

assert h.shape == (n, d) # Output

\end{lstlisting}

\subsubsection{Optional sharing of embedding matrices}\label{app:shared_embedding}

We use disjoint embeddings for the encoder and decoder by default, which allows us to use a different dimensionality for the encoder (typically kept at $d_{\textsc{Enc}} = 896)$ and for the decoder (that we scale up to $d=8192$). It is possible to share the embeddings, with little difference in training, as we show in the ablation section.

\subsection{Baseline to \retro model fine-tuning}
As shown in \autoref{fig:ft}, we found that we were able to take a pre-trained baseline transformer and add $\retro$ through fine-tuning.
In all cases, we froze all weights from pre-training and freshly initialised the retrieval encoder and cross-attention weights. In all cases, the cross-attention is added every third layer starting at layer six.
The learning rate for the three smaller models was set to $2 \times 10^{-4}$ and half that for the larger model.
We experimented with allowing the entire model to resume training during fine-tuning but consistently found that the best approach was to freeze the pre-trained model.
This kept the retrieval-off performance frozen whereas when all weights were tuned the retrieval off performance would degrade. 

\section{Training details and hyperparameters}

We provide the hyperparameters used in the various experiments of \autoref{sec-results}.

\subsection{Language model pre-training}
In \autoref{tab:model_hp}, we show the hyperparameters of the different models we train. In all cases, we train for 419,430,400,000 training tokens.
The three smaller models are trained with a batch size of 256 and the largest model is trained with a batch size of 1024. 
The minimum learning rate is set to 0.1 times the maximum learning rate, which is shown in \autoref{tab:model_hp}.
The learning rate is decayed using a cosine cycle length that matches the total number of training tokens.
All models are trained using \texttt{AdamW} \citep{loshchilov2018decoupled} with a weight decay parameter of 0.1.
The learning rate linearly increases from $10^{-7}$ to the maximum learning rate over the first 750 steps of training.
\begin{table}[h]
    \caption{\textbf{\retro model hyperparameters}, along with the size of the decoder.}
    
    \small
    \vspace{2.0mm}
    
    \centering
    \begin{tabular}{c c c c c c c c c c}
        Baseline & $d_{model}$ & $d_{ffw}$ & $\#$ heads & Head size & $\#$ layers & $P$ & $P_{\textsc{Enc}}$ & Max LR \\ 
        \toprule
        247M  & 896 & 3584 & 16 & 64 & 12 & $[6, 9, 12]$ & $[1]$ & 2$\times10^{-4}$   \\
        564M  & 1536 & 6144 & 12 & 128 & 12 & $[6, 9, 12]$ & $[1]$ & 2$\times10^{-4}$ \\
        1,574M & 2048 & 8192 & 16 & 128 & 24 & $[9, 12, \dots, 24]$ & $[1]$ & 2$\times10^{-4}$ \\
        7,505M & 4096 & 16384 & 32 & 128 & 32 & $[9, 12, \dots, 32]$ & $[1]$ & 1$\times10^{-4}$ \\
    \end{tabular}

    \label{tab:model_hp}
\end{table}
All models use ZeRO to shard the optimiser state \citep{rajbhandari_zero_2020}.
Additional infrastructure details can be found in \cite{rae2021gopher}.

\begin{table}[h]
    \caption{Hyperparameters for the Wikitext103 experiments presented in \autoref{table-wikitext103}. We use the same learning rate schedule for the baseline and the $\retro$-fitting. For $\retro$-fitting, we reset the schedule i.e. the schedule starts from step 0, not from step 35,000. }
    
    \small
    \vspace{2.0mm}

    \centering
    \begin{tabular}{l l r} 
    \toprule
      Model & Number of layers & 18 \\
      & $d$ & 1024 \\
      & $d_\ffw$ & 4096 \\ 
      & Key size & 64 \\
      & Value size & 64 \\
      & Number of heads & 16 \\ 

      \midrule
     Training data & Dataset & Wikitext103train \\
      & Sequence length &  3072 \\
      & Batch size & 128 \\
      & Tokenizer vocabulary size & 128,000 \\ \midrule
      Optimisation & optimiser & Adam \\
      & Adam's $\beta_1 $ & 0.9 \\
       & Adam's $\beta_2$ & 0.95 \\
       & Adam's $\epsilon$ & 1e-8 \\
       & Dropout rate & 0.25 \\ \hline
      Schedule  & Learning rate start & 1e-7 \\
       & Learning rate max & 2.5e-4 \\
       & Learning rate min & 2e-5 \\
       & Warmup steps & 4,000 \\
       & Cosine cycle steps & 100,000 \\ \hline
      Evaluation & Overlapping proportion & 87.5 \% \\ 
      \bottomrule
      
    \end{tabular}

    \label{tab:app-wikitext103-hyper}

\end{table}

\subsection{Wikitext103 comparison}
\label{subsubsec-app-wikitext103}
We provide more details on our Wikitext103 results presented in \autoref{subsec-scaling} and \autoref{table-wikitext103}. We train a baseline transformer on the Wikitext103 training set with the hyperparameters presented in \autoref{tab:app-wikitext103-hyper}. The learning rate ramps linearly from $1 \times 10^{-7}$ to $2.5 \times 10^{-4}$ in the first 4,000 steps, then decays to $2 \times 10^{-5}$ at 100,000 steps using a cosine schedule. The baseline checkpoint at step 35,000 has the lowest perplexity on Wikitext103 valid, of $21.58$, for overlapping proportion of 75\% (sliding window evaluation that only uses probabilities for tokens that have at least 75\% of the sequence length of context, when available). We use this checkpoint for all our baseline and $\knnlm$ numbers reported in \autoref{table-wikitext103}, except that \autoref{table-wikitext103} reports for an overlapping proportion of 87.5 \%, which slightly lowers the perplexity of our baseline to 21.53 on Wikitext103 valid.

\begin{figure}[h]
    \centering
    \includegraphics[width=0.99  \textwidth]{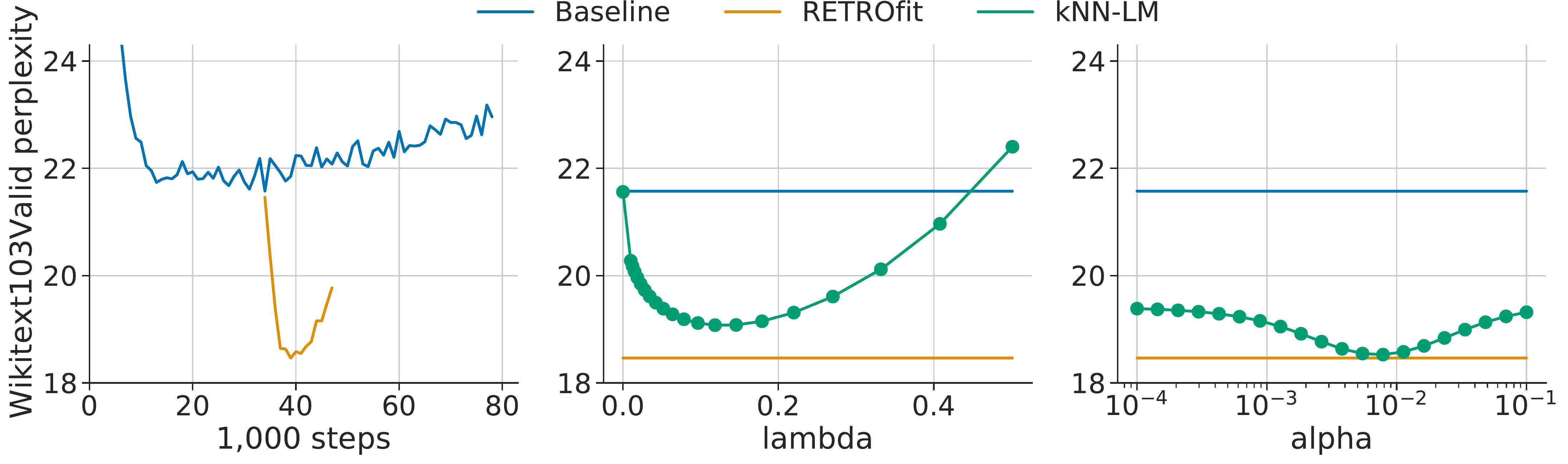}
   
    \caption{\textbf{Wikitext103valid perplexities.}  \textit{Left:} Baseline and $\retro$fit (initialized from baseline's checkpoint at 35,000 steps) perplexities as a function of training steps. \textit{Center and right:} $\knnlm$ perplexity as a function of $\lambda$ (for 
    $\alpha = 0.0012$) and $\alpha$ (for $\lambda =  0.12$) respectively.}

    \label{fig:app_wikitraining_curves}
\end{figure}

We also use the 35,000 step baseline checkpoint as initialization for a $\retro$fit, which otherwise uses the same optimiser and schedule hyperparameters but only trains the new retrieval weights, as explained in \autoref{subsec:ft}. Our best $\retro$fit checkpoint has a Wikitext103 valid perplexity $18.46$, when retrieving from Wikipedia. We use this $\retro$ checkpoint in \autoref{table-wikitext103} for all other retrieval sets. The evaluation curves for our baseline and $\retro$fit is shown if \autoref{fig:app_wikitraining_curves} (left). In this particular case, because Wikitext103 is quite small, training a $\retro$ model from scratch led to weaker results than the baseline, at least when retrieving from Wikipedia, as we couldn't find an effective way to mitigate the increased over-fitting due to the additional weights of $\retro$.

We also re-implement $\knnlm$ using the same tokenizer and dataset that we use for our baseline and $\retro$fitting experiments. $\knnlm$ has probabilities $p_{\knnlm}= \lambda p_{LM}  + (1-\lambda) p_{kNN}$ with $p_{kNN}(n_k) \propto \exp(-\alpha d_k)$. To tune $\lambda$ and $\alpha$, we begin with $\alpha=0.0012$, which corresponds to the inverse of the standard deviation of the norm of the embeddings that we use as keys and queries for $\knnlm$. We find the best $\lambda=0.118$. We then find the best $\alpha=0.00785$ for that value of $\lambda$. \autoref{fig:app_wikitraining_curves} center and right respectively show the perplexity of $\knnlm$ as a function of $\lambda$ and $\alpha$.

\subsection{\retrofitting baseline models experiments}
In \autoref{tab:app-retrofitting-hyper}, we give the hyperparameters used for \retrofitting the models on \MassiveText.
\label{appendix:retrofitting-details}
\begin{table}[h]
    \caption{Hyperparameters for the \retrofitting experiments }
    
    \small
    \vspace{2.0mm}

    \centering
    \begin{tabular}{l c c c} 
    \toprule
    Model & Layers with $\retro$-block ($P$)  & Learning rate & Batch size \\
    \midrule
   172M & Every 3\textsuperscript{rd} from 6 & $2 \times 10^{-4} \rightarrow 2 \times 10^{-5}$ & 256 \\
   425M & Every 3\textsuperscript{rd} from 6 & $2 \times 10^{-4} \rightarrow 2 \times 10^{-5}$ & 256 \\
   1.5B & Every 3\textsuperscript{rd} from 6 & $2 \times 10^{-4} \rightarrow 2 \times 10^{-5}$ & 256 \\
   7.5B & Every 3\textsuperscript{rd} from 6 &  $1 \times 10^{-4} \rightarrow 1 \times 10^{-5}$ & 256\\
    \bottomrule
      
    \end{tabular}

    \label{tab:app-retrofitting-hyper}

\end{table}

\subsection{Question answering experiments}
\label{appendix:natural-questions}

We fine-tune our 7.5B \retro model for 25,000 steps, using a batch size of 128, a learning rate cosine scheduled from $10^{-6}$ to $10^{-7}$, with a linear ramp of 750 steps. We use dropout in the decoder only, as it performs better than using dropout in both the encoder and the decoder. Each neighbour is formatted as \texttt{title: \{title\}, source: \{source\}}. We use the top 20 neighbours from $\dpr$ when training and evaluating.

\begin{table}[h]
    \caption{\textbf{Performance of \retro for different variants.} Model performance on C4 evaluation set, measured in bytes-per-bits, for a 247M parameter model trained with a 157 billion token schedule.}
    \label{table:ablation}
\small
    \centering
\begin{tabular}{llr}
\toprule
               Ablation group &                        Ablation &    C4 eval bpb \\
\midrule
                    Model &                      \retro &  0.822 \\
                     &       No query conditioning &  0.829 \\
                     &  No CA positional encodings &  0.826 \\
                     &           Shared embeddings &  0.823 \\
                     &             6-layer encoder &  0.821 \\
\midrule                     
         Retrieval values &                Neighbours N &  0.950 \\
          &             Continuations F &  0.895 \\
          &                No retrieval &  0.987 \\
\midrule
       Training neighbours &       1 training neighbours &  0.858 \\
        &       4 training neighbours &  0.847 \\
\midrule
 Cross attention position &         CA top layer (1/12) &  0.827 \\
  &         CA mid layer (6/12) &  0.823 \\
  &        CA top layer (12/12) &  0.831 \\
  &               CA all layers &  0.860 \\
  &           CA every 3 from 1 &  0.823 \\
\bottomrule
\end{tabular}
\end{table}

\section{Model ablations}\label{app:ablation}
\begin{figure}[h]
    \centering
    \includegraphics[width=\textwidth]{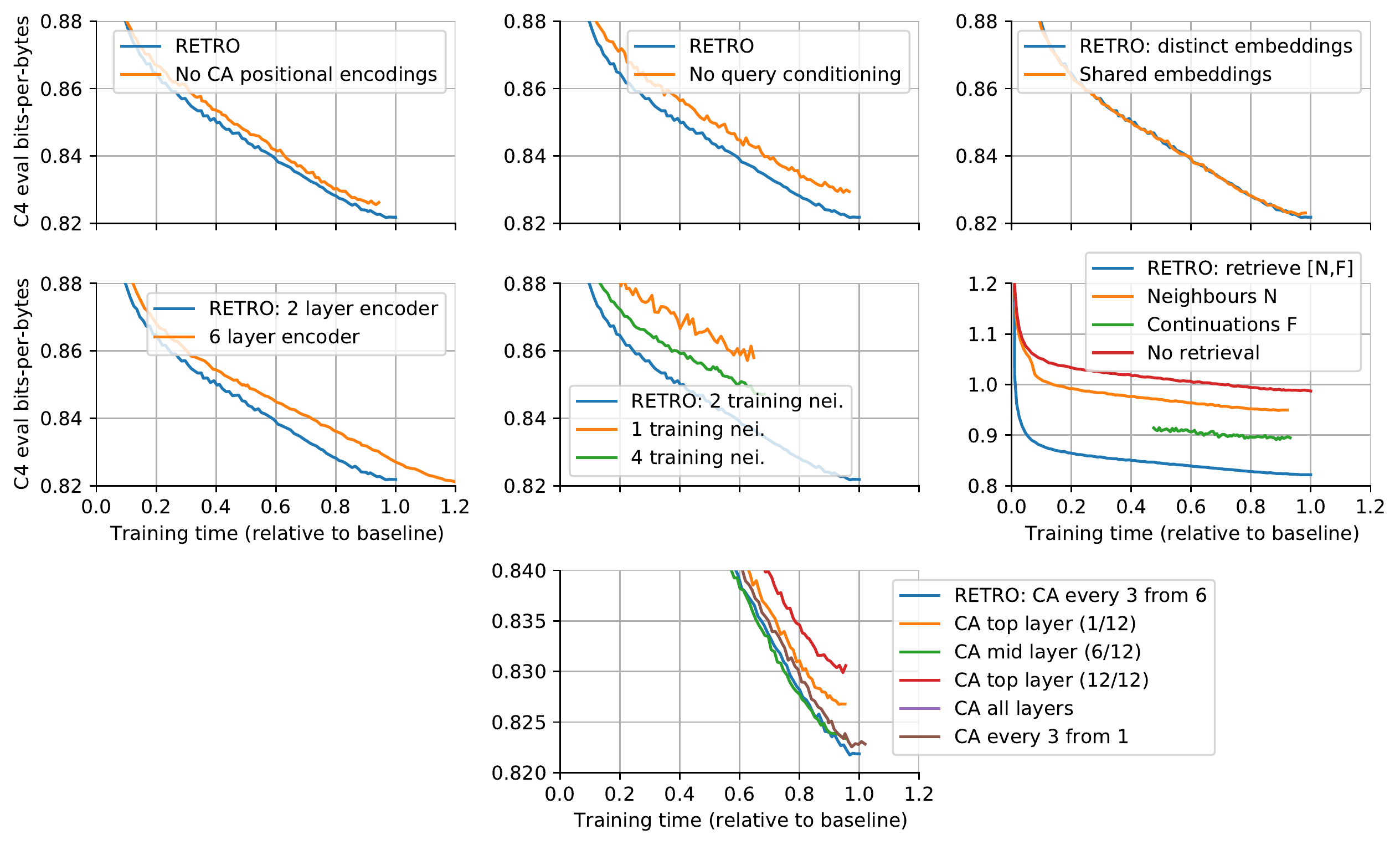}
    \caption{\textbf{Computational efficiency for different variants.} We report the training curves plotting C4 evaluation bytes per bits against time, relative to the time taken to train the baseline \retro model. Overall, our design choices are optimal in term of computational efficiency.}
    \label{fig:ablation}
\end{figure}

We validate important design choices by evaluating what happens when we do not include them.
We use the 247M parameter model for all experiments and we train on a compressed 157 billion token schedule for all ablation experiments.
We describe results relative to the default settings presented in the main text and recalled here. We report C4 evaluation loss at the end of the training process, and also compares how the evaluation loss decrease versus the training time, measured relatively to the baseline training time. Results are reported in \autoref{fig:ablation} and \autoref{table:ablation}.

\paragraph{Using relative encodings in cross-attention.} Using relative encodings in cross-attention, as described in \autoref{app:relative_pos}, provides a pure improvement both in the number of steps to reach a given performance and computational efficiency.

\paragraph{Conditioning the encoder on the previous chunk.} Conditioning the encoder on the previous chunk's intermediate embeddings, as described in \autoref{app:feed_forward}, provides a pure improvement both in term of number of steps and computational efficiency.

\paragraph{Sharing embeddings.} Sharing embeddings across the encoder and the decoder does not affect performance. This motivates us using separate embeddings, as it allows to have a narrower encoder than decoder as we scale up the decoder size.

\paragraph{Attending neighbours and their continuation.} \retro models are trained by attending, for a given chunk, to both the neighbours of the preceding chunk and their continuation in time. We measure how training and evaluating \retro models on neighbours only and their continuation only affects performance. Overall, attending to neighbours only provides $22\%$ of the performance improvement due to retrieval in \retro, while attending the future of the neighbours gives $56\%$ of the performance. Attending to both neighbours and their continuation is the most efficient choice both in term of final performance and training efficiency.\label{app:future_helps}

\paragraph{Training a deeper encoder.}
All models in the text use a relatively small \retro encoder.
We experimented with a $3\times$ deeper encoder. We found that this resulted in a tiny decrease in loss-- 0.15\% at the cost of a larger training time ($+20\%$). Overall, using a shallow encoder is the best choice in term of training efficiency.

\paragraph{Training with multiple neighbours.} We measure the effect of training on a single retrieved neighbour, as well as training on 4 neighbours (\retro uses 2 neighbours in training). Training on a single neighbour results in a large decrease in performance, while training on 4 neighbours does not give substantial performance improvement at the end of training, but induces a large computational overhead. Overall, we find that using 2 neighbours is the best choice in term of training efficiency. Furthermore, evaluation can be done with additional neighbours.

\paragraph{Frequency of cross-attention.}
We measure how the frequency of cross-attention in the decoder affects performance. Overall, attending only once at the top or the bottom layer is a bad choice, while attending once on a mid-depth layer is relatively sound. We choose to have cross-attention every 3 layer as this provides a good trade-off between performance and run-time.

\section{Qualitative experiments}\label{app:qualitative}

We illustrate the usage of \retro models by looking at the perplexity of evaluation samples and by producing samples autoregressively.

\subsection{Inspecting neighbours and perplexities on evaluation data}
To build an intuition of what kind of information is leveraged by $\retro$ models, we suggest to have a closer look at a few evaluation documents and the corresponding retrieved data in Tables \ref{tab:data_wikisept21_greatcircle}, \ref{tab:data_wikisept21_allireland}, \ref{tab:data_wikisept21_paralympics} and \ref{tab:data_wiki103_radclif}. In these tables, the 4 rows corresponds to the first 4 chunks of the documents. The left-most column shows the chunk $C_u$ from the document being evaluated, where each token is coloured by the negative cross entropy loss difference $L_{\retro\textsc{[Off]}} - L_{\retro}$, a positive value, coloured in yellow, indicates that $\retro$ performs better when it has access to neighbours data. The second columns also shows the evaluated chunk $C_u$ but where each token $i$ is coloured by the length of the longest common prefix (LCP) with the preceding neighbours, i.e. the largest integer $j$ such that the prefix $(x_{i-j -1}, \dotsc, x_{i})$ also appears in $\ret(C_{u-1})$. Conversely, columns three and four show the first two neighbours and their continuation, respectively $[N^1_u, F^1_u]$ and $[N^2_u, F^2_u]$ coloured by LCP with subsequent chunk $C_{u+1}$.  LCP colouring helps to visually identify where the evaluated document overlaps the retrieved data. Note that the first chunk, $C_1$, in the second column is not coloured as it does not have any preceding neighbours to compute LCP with. Similarly, we do not show the neighbours of the fourth chunk, as these are not used to condition any of the first four chunks. 

Our qualitative analysis exhibits two major behaviors. 

Firstly, we observe that sometimes, specific facts in $C_u$ can be extracted from the preceding neighbours $\ret(C_{u-1})$ and that this can correspond to significant reduction in loss from the $\retro$ model for the corresponding tokens. Some examples of such behavior include the journal name \emph{Publishers Weekly} in \autoref{tab:data_wikisept21_greatcircle}, the football team name \emph{Tyrone} in \autoref{tab:data_wikisept21_allireland} or the event dates \emph{25 August to 6 September 2020} in \autoref{tab:data_wikisept21_paralympics}. In these three examples, the evaluated data consists of recent Wikipedia articles written in September 2021, after we built our retrieval dataset (see section \autoref{appendix:wikipedia-sept21-desc}). Yet, relevant information to predict this new data was available in the pre-existing retrieval data and the $\retro$ model seems to be able to correctly leverage it. 

On the other hand, we also observe that some of the evaluation data can partially leak in our training and retrieval data, despite the use of deduplication. $\retro$ can dramatically exploit such leakage. \autoref{tab:data_wiki103_radclif} illustrates this behavior, where the chunks $C_2$ and $C_3$ largely overlaps $\ret(C_{1})$ and  $\ret(C_{2})$ respectively, up to small formatting differences, which leads to much lower $\retro$ loss for all the corresponding tokens. 
\autoref{fig:scaling-wrt-leakage} shows that it is possible to quantify how much of the $\retro$ loss reduction is due to each of these two behaviors, by filtering out evaluation chunks that overlaps with the retrieval set.

\subsection{Inspecting samples}

We can follow the same procedure as above on samples generated using \retro models, in order to better understand where retrieval data had an influence on sampling.
We show examples of samples obtained using the 7.5B  $\retro$ model  in \autoref{tab:beaver}, \ref{tab:sample_hamlet}, \ref{tab:sample_human_right} and \ref{tab:sample_pi}.

\subsection{Neighbour quantification}
\begin{figure}[ht]
    \centering
    \includegraphics[width=0.5  \textwidth]{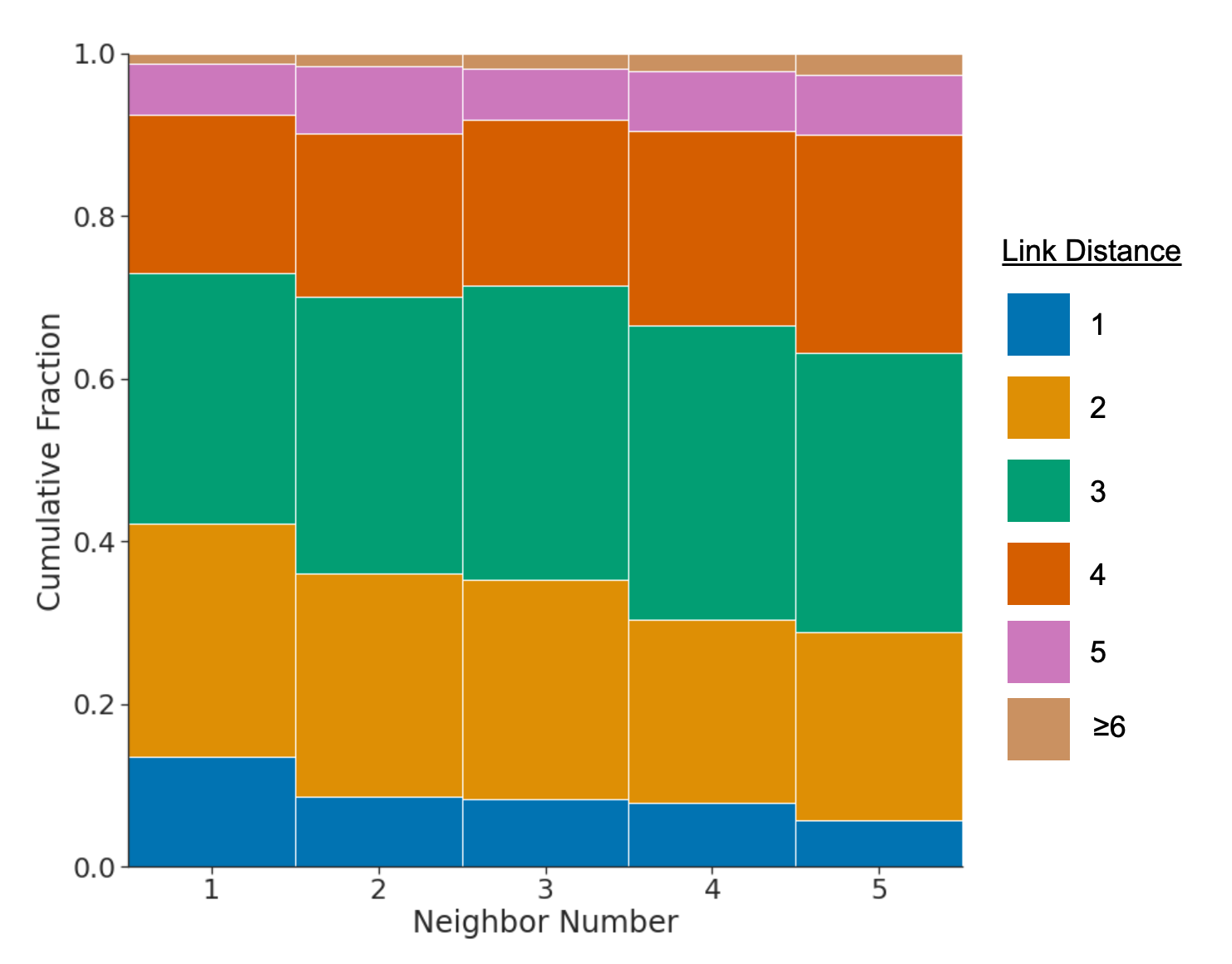}
    \caption{\textbf{Wikipedia link-distance between retrieved articles.} For each sequences, chunk combination we compute the link distance between the target and the top-5 neighbours using only Wikipedia. The rank shows the relative neighbour distance, where rank-1 is the first neighbour and rank 5 is the fifth. The different colours represent link distance. Because we do not retrieve from the same document, 1 is the smallest value.
    We find, on average, the distance between random articles with a path between them is over 5.0}
    \label{fig:supp_wiki}
\end{figure}
To quantify a notion of distance between the source document and the retrieved chunks, we can ask the distance between source articles when retrieving only from Wikipedia.
\citet{consonni2019wikilinkgraphs} provides a Wikipedia link dataset which, for each article, contains a list of neighbouring articles. Using this, we construct a directed graph and compute the distance  from one page to another.
In \autoref{fig:supp_wiki} we compute the link-distance between training sequences and the retrieved neighbours.
We find that retrieved documents tend to be from articles that are quite close to the article containing the target. 
Furthermore, we find that on average the distance increases with rank, suggesting that our neighbours are both useful and that the order is reasonable. 
This provides confidence for our larger-scale experiments where document distance is less well defined.

\section{Complementary quantitative results}

We report tables corresponding to quantitative figures of the main text, as well as further filtered language model results on the Pile.

\subsection{Main text datasets}

We report the performance of $\retro$ and baseline models, measured in bits-per-bytes on evaluation set, in
\autoref{tab:appendix-full-results}.

\begin{table}[h]
\caption{Full results for the main language modelling datasets. First three sets of rows correspond to \autoref{fig:summary}, last set of rows to \autoref{fig:scaling-wrt-params}.}
\label{tab:appendix-full-results}
\resizebox{\linewidth}{!}{
    \begin{tabular}{l |  c  c  c  c |  c  c  c  c |  c  c  c  c }
        & \multicolumn{4}{c|}{Baseline} & \multicolumn{4}{c|}{$\retro$ [Off]} &  \multicolumn{4}{c}{$\retro$[On]} \\
        & 172M & 425M & 1.5B & 7.5B & 172M & 425M & 1.5B & 7.5B & 172M & 425M & 1.5B & 7.5B \\
        \toprule
        C4 Eval bpb & 
        0.98 & 0.92 & 0.84 & 0.78 &
        0.98 & 0.92 & 0.84 & 0.78 &
        0.82 & 0.77 & 0.71 & 0.66 \\
        
        \midrule
        C4 Eval bpb (900B) &
        - & - & - & - &
        - & - & - & - &
        0.88 & 0.83 & 0.76 & 0.71 \\
        
        C4 Eval bpb (360B) &
        - & - & - & - &
        - & - & - & - &
        0.92 & 0.87 & 0.80 & 0.74 \\
        
        C4 Eval bpb (180B) &
        - & - & - & - &
        - & - & - & - &
        0.94 & 0.89 & 0.81 & 0.75 \\
        
        C4 Eval bpb (90B) &
        - & - & - & - &
        - & - & - & - &
        0.95 & 0.89 & 0.82 & 0.76 \\
        
        C4 Eval bpb (36B) &
        - & - & - & - &
        - & - & - & - &
        0.96 & 0.90 & 0.83 & 0.77 \\
        
        C4 Eval bpb (18B) &
        - & - & - & - &
        - & - & - & - &
        0.96 & 0.91 & 0.83 & 0.77 \\
        
        C4 Eval bpb (9B) &
        - & - & - & - &
        - & - & - & - &
        0.96 & 0.91 & 0.83 & 0.77 \\
        
        C4 Eval bpb (4B) &
        - & - & - & - &
        - & - & - & - &
        0.97 & 0.91 & 0.84 & 0.78 \\
        
        C4 Eval bpb (2B) &
        - & - & - & - &
        - & - & - & - &
        0.97 & 0.91 & 0.84 & 0.78 \\
        
        \midrule
        C4 Eval bpb ($k={1}$) &
        - & - & - & - &
        - & - & - & - &
        0.84 & 0.79 & 0.73 & 0.67 \\
        
        C4 Eval bpb ($k={2}$) &
        - & - & - & - &
        - & - & - & - &
        0.83 & 0.78 & 0.72 & 0.67 \\
        
        C4 Eval bpb ($k={3}$) &
        - & - & - & - &
        - & - & - & - &
        0.82 & 0.78 & 0.71 & 0.66 \\
        
        C4 Eval bpb ($k={4}$) &
        - & - & - & - &
        - & - & - & - &
        0.82 & 0.77 & 0.71 & 0.66 \\
        
        C4 Eval bpb ($k={5}$) &
        - & - & - & - &
        - & - & - & - &
        0.82 & 0.77 & 0.71 & 0.66 \\
        
        C4 Eval bpb ($k={10}$) &
        - & - & - & - &
        - & - & - & - &
        0.82 & 0.77 & 0.71 & 0.66 \\
        
        C4 Eval bpb ($k={20}$) &
        - & - & - & - &
        - & - & - & - &
        0.82 & 0.77 & 0.71 & 0.66 \\
        
        C4 Eval bpb ($k={30}$) &
        - & - & - & - &
        - & - & - & - &
        0.82 & 0.77 & 0.71 & 0.65 \\
        
        C4 Eval bpb ($k={40}$) &
        - & - & - & - &
        - & - & - & - &
        0.83 & 0.77 & 0.71 & 0.65 \\
        
        C4 Eval bpb ($k={50}$) &
        - & - & - & - &
        - & - & - & - &
        0.83 & 0.78 & 0.71 & 0.66 \\
        
        C4 Eval bpb ($k={60}$) &
        - & - & - & - &
        - & - & - & - &
        0.84 & 0.78 & 0.72 & 0.66 \\
        
        C4 Eval bpb ($k={70}$) &
        - & - & - & - &
        - & - & - & - &
        0.84 & 0.79 & 0.72 & 0.66 \\
        
        C4 Eval bpb ($k={80}$) &
        - & - & - & - &
        - & - & - & - &
        0.85 & 0.79 & 0.73 & 0.66 \\
        
        C4 Eval bpb ($k={90}$) &
        - & - & - & - &
        - & - & - & - &
        0.85 & 0.79 & 0.73 & 0.66 \\
        
        C4 Eval bpb ($k={100}$) &
        - & - & - & - &
        - & - & - & - &
        0.85 & 0.79 & - & 0.67 \\

        \midrule
        
        Lambada Accuracy &
        0.42 & 0.51 & 0.61 & 0.69 &
        0.47 & 0.54 & 0.63 & 0.70 &
        0.52 & 0.60 & 0.67 & 0.73 \\
        
        Curation Corpus bpb &
        0.69 & 0.63 & 0.56 & 0.52 &
        0.68 & 0.64 & 0.57 & 0.51 &
        0.66 & 0.61 & 0.55 & 0.50 \\

        Wikitext103 Perplexity  &
        25.62 & 19.29 & 13.98 & 10.65 &
        25.88 & 19.78 & 13.89 & 10.40 &
        3.32 & 2.96 & 2.53 & 2.22 \\

        Wikipedia Sept. 2021 bpb &
        0.85 & 0.78 & 0.71 & 0.65 &
        0.86 & 0.79 & 0.71 & 0.65 &
        0.79 & 0.73 & 0.66 & 0.61 \\

    \end{tabular}

}
    \vspace{2.0mm}
    \centering
\end{table}

\subsection{The Pile}

In \autoref{fig:pile_results}, we compare \retro against Jurassic-1 \citep{jurassic}. The full bits-per-bytes results are reported in \autoref{tab:appendix-pile}.
\begin{table}[h]
    \caption{\textbf{Full results on The Pile, measured in bits-per-bytes.} Jurassic-1 and GPT-3 numbers are taken from \citet{jurassic}. Gopher numbers are taken from \citet{rae2021gopher}.}
    \small
    \vspace{2.0mm}
    \centering
    \begin{tabular}{l  c  c  c  c c }
        Subset & 7B Baseline (Ours) & GPT-3 & Jurassic-1 & Gopher & 7.5B $\retro$\\ 
        \toprule
        arxiv & 0.742 & 0.838 & 0.680 & \textbf{0.641} & 0.714 \\
        books3 & 0.792 & 0.802 & 0.835 & 0.706 & \textbf{0.653} \\
        dm\_mathematics & 1.177 & 1.371 & \textbf{1.037} & 1.135 & 1.164 \\
        freelaw & 0.576 & 0.612 & 0.514 & 0.506 & \textbf{0.499} \\
        github & 0.420 & 0.645 & 0.358 & 0.367 & \textbf{0.199} \\
        gutenberg\_pg\_19 & 0.803 & 1.163 & 0.890 & 0.652 & \textbf{0.400} \\
        hackernews & 0.971 & 0.975 & 0.869 & 0.888 & \textbf{0.860} \\
        nih\_exporter & 0.650 & 0.612 & \textbf{0.590} & 0.590 & 0.635 \\
        opensubtitles & 0.974 & 0.932 & \textbf{0.879} & 0.894 & 0.930 \\
        philpapers & 0.760 & 0.723 & 0.742 & \textbf{0.682} & 0.699 \\
        pile\_cc & 0.771 & 0.698 & 0.669 & 0.688 & \textbf{0.626} \\
        pubmed\_abstracts & 0.639 & 0.625 & 0.587 & 0.578 & \textbf{0.542} \\
        pubmed\_central & 0.588 & 0.690 & 0.579 & 0.512 & \textbf{0.419} \\
        stackexchange & 0.714 & 0.773 & 0.655 & 0.638 & \textbf{0.624} \\
        ubuntu\_irc & 1.200 & 0.946 & \textbf{0.857} & 1.081 & 1.178 \\
        uspto\_backgrounds & 0.603 & 0.566 & \textbf{0.537} & 0.545 & 0.583 \\
    \end{tabular}

    \label{tab:appendix-pile}
\end{table}

\subsection{Filtered results}\label{app:leakage}

\begin{figure}[h]
    \centering
    \includegraphics[width=\textwidth]{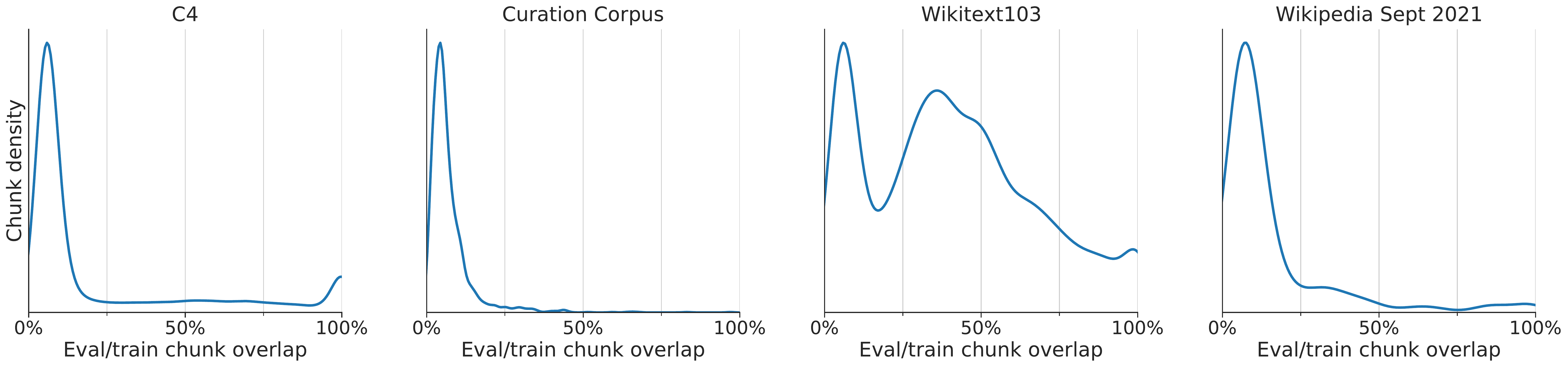}
    \caption{\textbf{Distribution of the overlap between evaluation and train chunks} for C4, Curation Corpus, Wikitext103 and Wikipedia Sept. 2021.}
    \label{fig:leakage_distribution_main}
\end{figure}

\paragraph{Distribution of leaked chunks in our main evaluation sets.} We evaluate leakage between the evaluation sets and the training set by measuring the proportion of evaluation chunks with a certain overlap $r(C)$. We show histograms in \autoref{fig:leakage_distribution_main}. We can see that $C4$ has some slight overlaps between train and evaluation. Similarly, chunks of Wikitext103 appear in the training set despite having removed the actual Wikitext103 evaluation documents from the training set.
On the other hand, our Wikipedia September 21 dataset shows almost no leakage (data being original documents that did not exist at training data creation), and neither does Curation Corpus.

\begin{figure}[t]
    \centering
    \includegraphics[width=\textwidth]{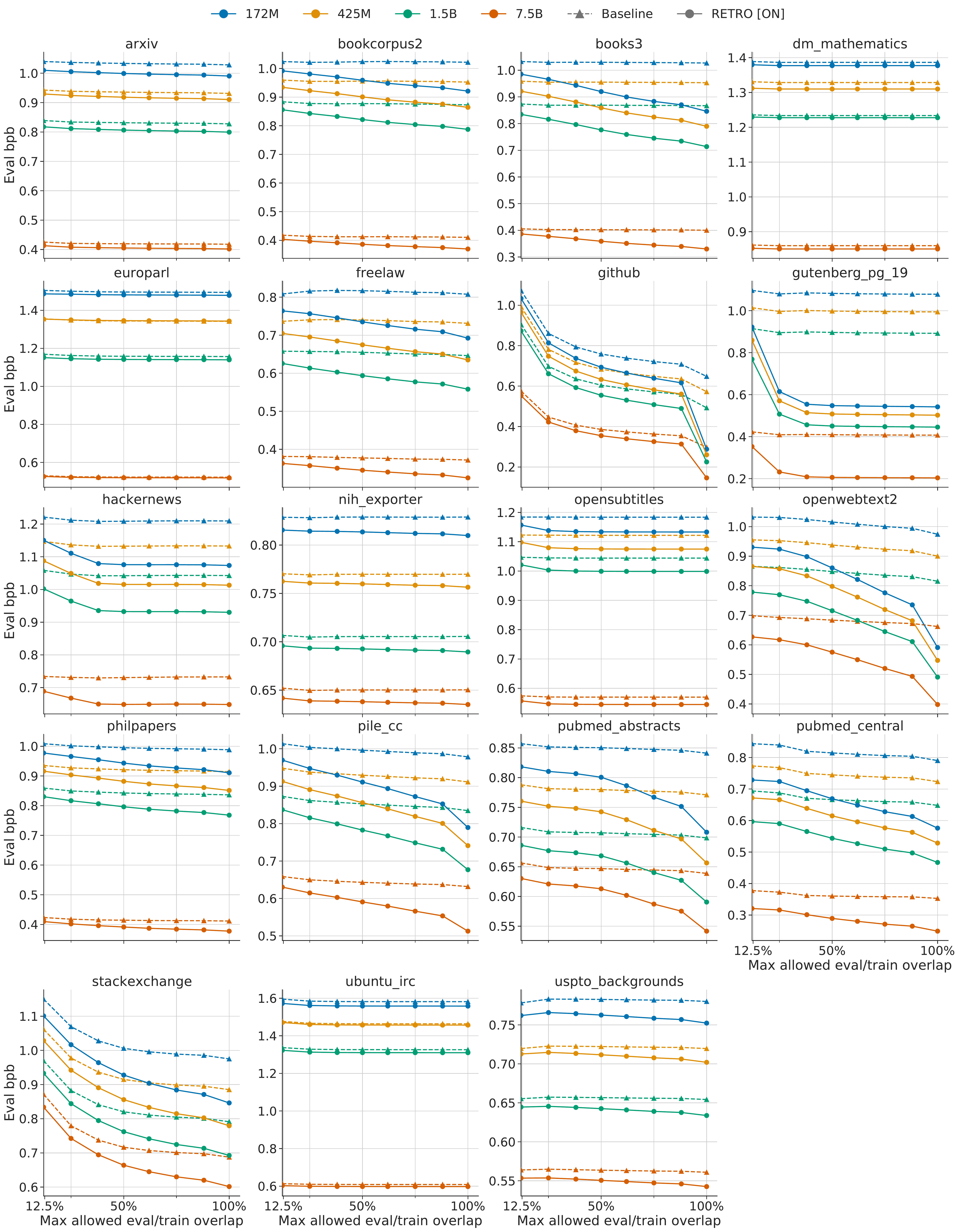}
    \caption{{\bf Filtered evaluation losses on the Pile}, with baseline Transformers and \retro.}
    \label{fig:leakage_pile}
\end{figure}

\begin{figure}[t]
    \centering
    \includegraphics[width=\textwidth]{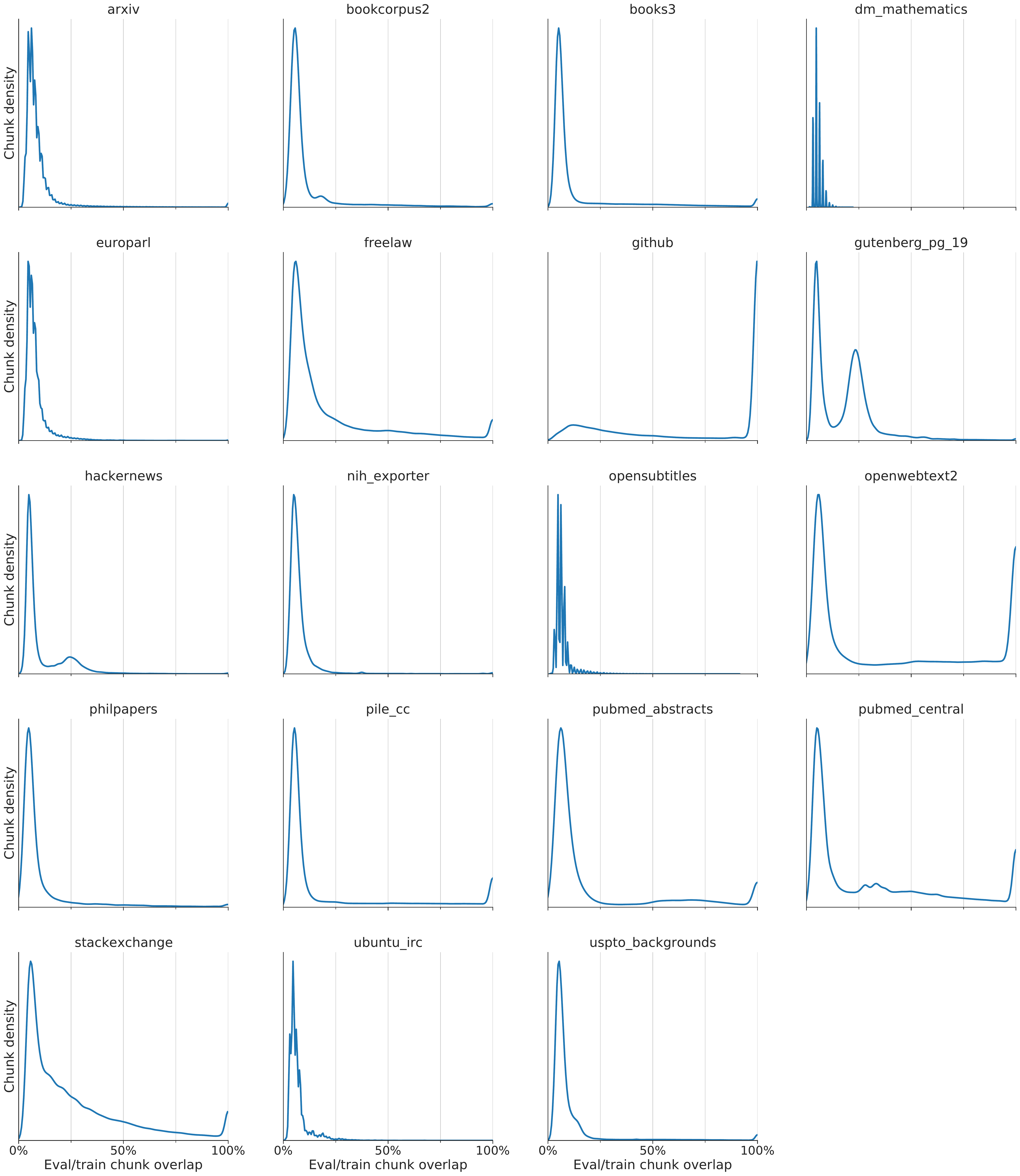}
    \caption{{\bf Distribution of the overlap between evaluation and train chunks} for the Pile evaluation sets.}
    \label{fig:distribution_pile}
\end{figure}

\paragraph{Filtered results on the Pile.} We report chunk overlap distribution and filtered performance curves on the Pile in \autoref{fig:distribution_pile} and \autoref{fig:leakage_pile}, respectively.
The qualitative interpretation of the filtered curves is the same: \retro models exploit leakage more, but the performance improvement they provide remains significant even on original chunks that haven't been observed in the training set.

\begin{table}[h]     \tiny      \caption{\small \textbf{Great Circle (novel)}, from Wikipedia September 21. The article is about a recent novel and chunks $C_3$ and $C_4$ are specifically about its reception. The name \textbf{Publishers Weekly} of the journal that reviewed the novel appears both in the neighbours $[N^1_3, F^1_3],[N^2_3, F^2_3]$ of chunk $C_3$ and  in the subsequent chunk $C_4$, where the loss for those tokens is significantly reduced by $\retro$. }  
   \vspace{2.0mm}       \centering     
\begin{tabular}{ l l l l }
\datavisheader
\colorbox{cm_32!40}{\strut Great\ Circle\ (novel)Great\ Circle\ i} & Great\ Circle\ (novel)  Great\ Circle\ i & \colorbox{cm_0!40}{\strut The\ Dutch\ House\ (novel)The\ Dutch\ H} & \colorbox{cm_0!40}{\strut The\ Dutch\ House\ (novel)The\ Dutch\ H} \\ 
 \colorbox{cm_32!40}{\strut s\ a\ 2021\ novel\ by\ Maggie\ Shipstead,\ } & s\ a\ 2021\ novel\ by\ Maggie\ Shipstead,\  & \colorbox{cm_0!40}{\strut ouse}\colorbox{cm_12!40}{\strut \ is\ a}\colorbox{cm_0!40}{\strut \ 2019\ novel\ by\ Ann\ Patchett} & \colorbox{cm_0!40}{\strut ouse}\colorbox{cm_12!40}{\strut \ is\ a}\colorbox{cm_0!40}{\strut \ 2019\ novel\ by\ Ann\ Patchett} \\ 
 \colorbox{cm_32!40}{\strut published\ on\ May\ 4,\ 2021,\ by\ Alfred\ } & published\ on\ May\ 4,\ 2021,\ by\ Alfred\  & \colorbox{cm_12!40}{\strut .}\colorbox{cm_0!40}{\strut \ It\ was\ published\ by\ Harper\ on\ Sept} & \colorbox{cm_12!40}{\strut .}\colorbox{cm_0!40}{\strut \ It\ was\ published\ by\ Harper\ on\ Sept} \\ 
 \colorbox{cm_32!40}{\strut A.\ Knopf.The\ novel\ has\ been\ shortl} & A.\ Knopf.  The\ novel\ has\ been\ shortl & \colorbox{cm_0!40}{\strut ember\ 24,\ 2019.\ It\ tells}\colorbox{cm_12!40}{\strut \ the}\colorbox{cm_0!40}{\strut \ story\ o} & \colorbox{cm_0!40}{\strut ember\ 24,\ 2019.\ It\ tells}\colorbox{cm_12!40}{\strut \ the}\colorbox{cm_0!40}{\strut \ story\ o} \\ 
 \colorbox{cm_32!40}{\strut isted\ for\ the\ 2021\ Booker\ Prize.Sy} & isted\ for\ the\ 2021\ Booker\ Prize.  Sy & \colorbox{cm_0!40}{\strut f}\colorbox{cm_12!40}{\strut \ a}\colorbox{cm_0!40}{\strut \ brother\ and\ sister\ over}\colorbox{cm_12!40}{\strut \ the}\colorbox{cm_0!40}{\strut \ cour} & \colorbox{cm_0!40}{\strut f}\colorbox{cm_12!40}{\strut \ a}\colorbox{cm_0!40}{\strut \ brother\ and\ sister\ over}\colorbox{cm_12!40}{\strut \ the}\colorbox{cm_0!40}{\strut \ cour} \\
\bottomrule 
  \end{tabular}
    \label{tab:data_wikisept21_greatcircle}
\end{table}
   \begin{table}[h]       \tiny    
    \caption{\small \textbf{All-Ireland Senior Football Championship Final}, from Wikipedia September 21. The name of the team \textbf{Tyrone}  appears both in the second neighbours $[N^2_1, F^2_1]$ of chunk $C_1$ and  in the subsequent chunk $C_2$, where the loss for those tokens is significantly reduced by $\retro$. }  
 
      \vspace{2.0mm}       \centering     \begin{tabular}{ l l l l }
\datavisheader

 \colorbox{cm_32!40}{\strut 2021\ All-Ireland\ Senior\ Football\ Cha} & 2021\ All-Ireland\ Senior\ Football\ Cha & \colorbox{cm_0!40}{\strut 2018\ All}\colorbox{cm_12!40}{\strut -}\colorbox{cm_0!40}{\strut Ireland\ Senior\ Football\ Cha} & \colorbox{cm_0!40}{\strut 2018\ All}\colorbox{cm_12!40}{\strut -}\colorbox{cm_0!40}{\strut Ireland\ Senior\ Football\ Cha} \\ 
 \colorbox{cm_32!40}{\strut mpionship\ FinalThe\ 2021\ All-Irelan} & mpionship\ Final  The\ 2021\ All-Irelan & \colorbox{cm_0!40}{\strut mpionship\ FinalThe\ 2018\ All}\colorbox{cm_12!40}{\strut -}\colorbox{cm_0!40}{\strut Irelan} & \colorbox{cm_0!40}{\strut mpionship\ FinalThe\ 2018\ All}\colorbox{cm_12!40}{\strut -}\colorbox{cm_0!40}{\strut Irelan} \\ 
 \colorbox{cm_32!40}{\strut d\ Senior\ Football\ Championship\ Final} & d\ Senior\ Football\ Championship\ Final & \colorbox{cm_0!40}{\strut d\ Senior\ Football\ Championship\ Final} & \colorbox{cm_0!40}{\strut d\ Senior\ Football\ Championship\ Final} \\ 
 \colorbox{cm_32!40}{\strut \ was\ the\ 134th\ final\ of\ the\ All-Irel} & \ was\ the\ 134th\ final\ of\ the\ All-Irel & \colorbox{cm_12!40}{\strut \ was\ the}\colorbox{cm_0!40}{\strut \ 131st}\colorbox{cm_12!40}{\strut \ final}\colorbox{cm_0!40}{\strut \ of}\colorbox{cm_12!40}{\strut \ the}\colorbox{cm_0!40}{\strut \ All}\colorbox{cm_12!40}{\strut -}\colorbox{cm_0!40}{\strut Irel} & \colorbox{cm_12!40}{\strut \ was\ the}\colorbox{cm_0!40}{\strut \ 131st}\colorbox{cm_12!40}{\strut \ final}\colorbox{cm_0!40}{\strut \ of}\colorbox{cm_12!40}{\strut \ the}\colorbox{cm_0!40}{\strut \ All}\colorbox{cm_12!40}{\strut -}\colorbox{cm_0!40}{\strut Irel} \\ 
 \colorbox{cm_32!40}{\strut and\ Senior\ Football\ Championship\ and} & and\ Senior\ Football\ Championship\ and & \colorbox{cm_0!40}{\strut and\ Senior\ Football\ Championship\ and} & \colorbox{cm_0!40}{\strut and\ Senior\ Football\ Championship\ and} \\
\bottomrule
    \end{tabular}
      \label{tab:data_wikisept21_allireland}
  \end{table}
   \begin{table}[h]       \tiny    
    \caption{\small \textbf{2020 Summer Paralympics}, from Wikipedia September 21. The original dates of the event, \textbf{25 August to 6 September 2020},  appears both in the neighbors $[N^1_1, F^1_1], [N^2_1, F^2_1]$ of chunk $C_1$ and  in the subsequent chunk $C_2$, where the loss for those tokens is significantly reduced by $\retro$. Interestingly, in this case, the neighbors were written at a time when the event hadn't yet been postponed.}  
 
      \vspace{2.0mm}       \centering     \begin{tabular}{ l l l l }
\datavisheader
   
   \colorbox{cm_32!40}{\strut 2020\ Summer\ ParalympicsThe\ ,\ brand} & 2020\ Summer\ Paralympics  The\ ,\ brand & \colorbox{cm_12!40}{\strut p}\colorbox{cm_25!40}{\strut ics}\colorbox{cm_12!40}{\strut \ Games.}\colorbox{cm_0!40}{\strut *\ The}\colorbox{cm_12!40}{\strut \ 2020}\colorbox{cm_25!40}{\strut \ Summer}\colorbox{cm_12!40}{\strut \ Paraly} & \colorbox{cm_0!40}{\strut 2020}\colorbox{cm_12!40}{\strut \ Summer\ Paralym}\colorbox{cm_25!40}{\strut p}\colorbox{cm_38!40}{\strut ics}\colorbox{cm_0!40}{\strut The}\colorbox{cm_12!40}{\strut \ }\colorbox{cm_0!40}{\strut \ are\ an} \\ 
 \colorbox{cm_32!40}{\strut ed\ as\ the\ Tokyo\ 2020\ Paralympic\ Game} & ed\ as\ the\ Tokyo\ 2020\ Paralympic\ Game & \colorbox{cm_12!40}{\strut m}\colorbox{cm_25!40}{\strut p}\colorbox{cm_38!40}{\strut ics}\colorbox{cm_0!40}{\strut \ are\ an\ upcoming\ major\ internat} & \colorbox{cm_0!40}{\strut \ upcoming\ major\ international\ multi-} \\ 
 \colorbox{cm_32!40}{\strut s,\ was\ an\ international\ multi-sport\ } & s,\ was\ an\ international\ multi-sport\  & \colorbox{cm_0!40}{\strut ional\ multi-sport\ event}\colorbox{cm_12!40}{\strut \ for}\colorbox{cm_0!40}{\strut \ athletes} & \colorbox{cm_0!40}{\strut sport\ event}\colorbox{cm_12!40}{\strut \ for}\colorbox{cm_0!40}{\strut \ athletes}\colorbox{cm_12!40}{\strut \ with}\colorbox{cm_0!40}{\strut \ disabi} \\ 
 \colorbox{cm_32!40}{\strut parasports\ event\ held\ from\ 24\ August} & parasports\ event\ held\ from\ 24\ August & \colorbox{cm_12!40}{\strut \ with}\colorbox{cm_0!40}{\strut \ disabilities\ governed}\colorbox{cm_12!40}{\strut \ by\ the}\colorbox{cm_0!40}{\strut \ I} & \colorbox{cm_0!40}{\strut lities\ governed}\colorbox{cm_12!40}{\strut \ by\ the}\colorbox{cm_0!40}{\strut \ International} \\
\hline 
    \end{tabular}
      \label{tab:data_wikisept21_paralympics}
  \end{table}
   \begin{table}[h]       \tiny    
    \caption{\small \textbf{Daniel Radcliffe}, from Wikitext103Valid, retrieval data from c4. The chunks $C_2$ and $C_3$ are almost entirely retrieved from neighbours $[N_1, F_1]$ and $[N_2, F_2]$ respectively, up to formatting differences, which dramatically reduces the loss for these tokens. This example illustrates that when training data leaks into evaluation sets despite deduplication, our $\retro$ model can directly exploit this leakage.}  
 
      \vspace{2.0mm}       \centering     \begin{tabular}{ l l l l }
\datavisheader

\colorbox{cm_32!40}{\strut =\ Daniel\ Radcliffe\ =Daniel\ Jacob\ R} & =\ Daniel\ Radcliffe\ =  Daniel\ Jacob\ R & \colorbox{cm_0!40}{\strut Daniel\ Jacob\ Rad}\colorbox{cm_12!40}{\strut cliffe}\colorbox{cm_0!40}{\strut \ (born\ 23\ July} & \colorbox{cm_0!40}{\strut Daniel\ Jacob\ Rad}\colorbox{cm_12!40}{\strut cliffe}\colorbox{cm_0!40}{\strut \ (born\ 23\ July} \\ 
 \colorbox{cm_32!40}{\strut adcliffe\ (\ born\ 23\ July\ 1989\ )\ is\ an} & adcliffe\ (\ born\ 23\ July\ 1989\ )\ is\ an & \colorbox{cm_0!40}{\strut \ 1989)\ is\ an\ English\ actor\ who\ rose}\colorbox{cm_12!40}{\strut \ } & \colorbox{cm_0!40}{\strut \ 1989)\ is\ an\ English\ actor\ who\ rose}\colorbox{cm_12!40}{\strut \ } \\ 
 \colorbox{cm_32!40}{\strut \ English\ actor\ who\ rose\ to\ prominenc} & \ English\ actor\ who\ rose\ to\ prominenc & \colorbox{cm_12!40}{\strut to}\colorbox{cm_0!40}{\strut \ prominence}\colorbox{cm_12!40}{\strut \ as\ the}\colorbox{cm_0!40}{\strut \ title\ character} & \colorbox{cm_12!40}{\strut to}\colorbox{cm_0!40}{\strut \ prominence}\colorbox{cm_12!40}{\strut \ as\ the}\colorbox{cm_0!40}{\strut \ title\ character} \\ 
 \colorbox{cm_32!40}{\strut e\ as\ the\ title\ character\ in\ the\ Harr} & e\ as\ the\ title\ character\ in\ the\ Harr & \colorbox{cm_12!40}{\strut \ in}\colorbox{cm_25!40}{\strut \ the}\colorbox{cm_12!40}{\strut \ Harry}\colorbox{cm_25!40}{\strut \ Potter}\colorbox{cm_38!40}{\strut \ film}\colorbox{cm_12!40}{\strut \ series.}\colorbox{cm_0!40}{\strut \ He} & \colorbox{cm_12!40}{\strut \ in}\colorbox{cm_25!40}{\strut \ the}\colorbox{cm_12!40}{\strut \ Harry}\colorbox{cm_25!40}{\strut \ Potter}\colorbox{cm_38!40}{\strut \ film}\colorbox{cm_12!40}{\strut \ series.}\colorbox{cm_0!40}{\strut \ He} \\
\bottomrule 
  \end{tabular}
    \label{tab:data_wiki103_radclif}
\end{table}

 \begin{table}[h]     \tiny    \caption{\small \textbf{Sample - Déclaration des droits de l'homme: Article premier.} The $\retrooff$ sample has correct syntax and is almost plausible but is hallucinated. The $\retroon$ sample is correctly copied from neighbour data, and robustly re-formated according to our prompt. }    \vspace{2.0mm}     \centering     \begin{tabular}{ l  l  l  l} \samplevisheader Article\ premier  -\ Les\ hommes\  & Article\ premier  -\ Les\ hommes\  & \colorbox{cm_12!40}{\strut \ de}\colorbox{cm_25!40}{\strut \ l}\colorbox{cm_0!40}{\strut ’}\colorbox{cm_12!40}{\strut homme\ et}\colorbox{cm_0!40}{\strut \ du\ citoyen}\colorbox{cm_12!40}{\strut .}\colorbox{cm_25!40}{\strut }\colorbox{cm_38!40}{\strut }\colorbox{cm_51!40}{\strut Ar} & \colorbox{cm_12!40}{\strut }\colorbox{cm_0!40}{\strut Les\ hommes\ naissent}\colorbox{cm_12!40}{\strut \ et}\colorbox{cm_0!40}{\strut \ demeur} \\ 
 naissent\ et\ demeurent\ libres\ e & naissent\ et\ demeurent\ libres\ e & \colorbox{cm_51!40}{\strut ticle}\colorbox{cm_0!40}{\strut \ 1er}\colorbox{cm_12!40}{\strut }\colorbox{cm_25!40}{\strut }\colorbox{cm_0!40}{\strut Les\ hommes\ naissent} & \colorbox{cm_0!40}{\strut ent\ libres}\colorbox{cm_12!40}{\strut \ et}\colorbox{cm_0!40}{\strut \ égaux\ en}\colorbox{cm_12!40}{\strut \ droits.} \\ 
 t\ égaux\ en\ droits.\ Les\ distinc & t\ égaux\ en\ droits.\ Les\ distinc & \colorbox{cm_12!40}{\strut \ et}\colorbox{cm_0!40}{\strut \ demeurent\ libres}\colorbox{cm_12!40}{\strut \ et}\colorbox{cm_0!40}{\strut \ égaux\ } & \colorbox{cm_0!40}{\strut \ Les\ distinctions\ sociales}\colorbox{cm_12!40}{\strut \ ne}\colorbox{cm_0!40}{\strut \ } \\ 
 tions\ sociales\ ne\ peuvent\ être & tions\ sociales\ ne\ peuvent\ être & \colorbox{cm_0!40}{\strut en}\colorbox{cm_12!40}{\strut \ droits.}\colorbox{cm_0!40}{\strut \ Les\ distinctions\ so} & \colorbox{cm_0!40}{\strut peuvent\ être\ fondées\ que\ sur}\colorbox{cm_12!40}{\strut \ l} \\
\end{tabular}
    \label{tab:sample_human_right}
\end{table}
\begin{table}[h]     \tiny      \caption{\small \textbf{Sample - Decimals of $\pi$.} The $\retrooff$ sample quickly diverges two digits after the end of the prompt whereas $\retroon$ correctly outputs a large number of $\pi$ digits, directly copied from the neighbours data.}     \vspace{2.0mm}     \centering     \begin{tabular}{ l  l  l  l} \samplevisheader Pi\ =\ 3. 1415926535\ 8979323846\  & Pi\ =\ 3. 1415926535\ 8979323846\  & \colorbox{cm_0!40}{\strut ``}\colorbox{cm_12!40}{\strut }\colorbox{cm_0!40}{\strut 141592}\colorbox{cm_12!40}{\strut 6535}\colorbox{cm_0!40}{\strut \ 8979323846\ 264}\colorbox{cm_12!40}{\strut 33} & \colorbox{cm_0!40}{\strut 46\ 264}\colorbox{cm_12!40}{\strut 33}\colorbox{cm_0!40}{\strut 8327}\colorbox{cm_12!40}{\strut 9\ 50}\colorbox{cm_0!40}{\strut 28841971}\colorbox{cm_12!40}{\strut \ 6}\colorbox{cm_0!40}{\strut 9399} \\ 
 2643383279\ 5028841971\ 69399375 & 2643383279\ 5028841971\ 69399375 & \colorbox{cm_0!40}{\strut 8327}\colorbox{cm_12!40}{\strut 9\ 50}\colorbox{cm_0!40}{\strut 28841971}\colorbox{cm_12!40}{\strut \ 6}\colorbox{cm_0!40}{\strut 9399375}\colorbox{cm_12!40}{\strut 1058} & \colorbox{cm_0!40}{\strut 375}\colorbox{cm_12!40}{\strut 10}\colorbox{cm_0!40}{\strut \ 582097}\colorbox{cm_12!40}{\strut 4944\ 5}\colorbox{cm_0!40}{\strut 92}\colorbox{cm_12!40}{\strut 30}\colorbox{cm_0!40}{\strut 78164}\colorbox{cm_12!40}{\strut }\colorbox{cm_0!40}{\strut 06} \\ 
 10 5820974944\ 5923078164\ 06286 & 10 5820974944\ 5923078164\ 06286 & \colorbox{cm_0!40}{\strut 2097}\colorbox{cm_12!40}{\strut 4944\ 5}\colorbox{cm_0!40}{\strut 92}\colorbox{cm_12!40}{\strut 30}\colorbox{cm_0!40}{\strut 78164\ 0628}\colorbox{cm_12!40}{\strut 6}\colorbox{cm_0!40}{\strut 20899} & \colorbox{cm_0!40}{\strut 28}\colorbox{cm_12!40}{\strut 6}\colorbox{cm_0!40}{\strut 20899\ 8628034}\colorbox{cm_12!40}{\strut 8}\colorbox{cm_0!40}{\strut 25\ 34}\colorbox{cm_12!40}{\strut 21}\colorbox{cm_0!40}{\strut 17067}\colorbox{cm_12!40}{\strut 9} \\
\end{tabular}
    \label{tab:sample_pi}
\end{table}
\end{document}